\documentclass[12pt]{amsart}
\usepackage[utf8]{inputenc}
\usepackage{amssymb,amsbsy,amsmath,amscd,amsthm,amsfonts,MnSymbol,mathrsfs}
\usepackage{cite}
\usepackage{natbib}
\usepackage[colorlinks]{hyperref}
\usepackage{graphicx}
\usepackage{verbatim}
\graphicspath{{Images/}}
\usepackage{float}

\usepackage{bm}
\usepackage{subfigure}

\newtheorem{definition}{Definition}
\newtheorem{theorem}{Theorem}
\newtheorem{example}{Example}

\newtheorem{proposition}{Proposition}

\newcommand{\cX}{\mathcal X}
\newcommand{\cH}{\mathcal H}

\newcommand{\cP}{\mathcal P}

\newcommand{\E}{\mathbb E}

\newcommand{\PP}{\mathbb{P}}

\newcommand{\RR}{\mathbb{R}}

\newcommand{\bN}{{\bf N}}
\newcommand{\bX}{{\bf X}}
\newcommand{\bZ}{{\bf Z}}

\newcommand{\bx}{{\bf x}}
\newcommand{\by}{{\bf y}}
\newcommand{\bz}{{\bf z}}
\newcommand{\ba}{{\bf a}}

\newcommand{\boldb}{{\bf b}}

\newcommand{\bs}{{\bf s}}
\newcommand{\bt}{{\bf t}}

\newcommand{\bv}{{\bf v}}

\newcommand{\bw}{{\bf w}}

\newcommand{\bomega}{{\bm \omega}}

\newcommand{\bvarepsilon}{{\bm \varepsilon}}

\newcommand{\br}{{\bf r}}

\newcommand{\BX}{{\bf X}}

\newcommand{\BY}{{\bf Y}}

\newcommand{\BZ}{{\bf Z}}

\newcommand{\BV}{{\bf V}}
\newcommand{\bS}{{\bf S}}

\newcommand{\0}{\mathbf{0}}

\newcommand{\one}{\mathbf{1}}

\newcommand{\bbP}{\mathbb{P}}
\newcommand{\bbr}{\mathbb{R}}

\newcommand{\bbS}{\mathbb{S}}

\newcommand{\calC}{\mathcal{C}}

\newcommand{\calI}{\mathcal{I}}

\newcommand{\calS}{\mathcal{S}}

\newcommand{\var}{\mathrm{Var}}

\newcommand{\cip}{\stackrel{P}{\to}}
\newcommand{\civ}{\stackrel{v}{\to}}

\newcommand{\bma}{\begin{matrix*}[r]}
\newcommand{\ema}{\end{matrix*}}

\newcommand{\vep}{\varepsilon}

\newcommand{\blue}{\textcolor{blue}}

\title{Kernel PCA  for multivariate extremes}

\author{Marco Avella Medina}
 \address{Department of Statistics \\
 	Columbia University}
 \email{marco.avella@columbia.edu}
 
 \author{Richard A. Davis}
 \address{Department of Statistics \\
 	Columbia University}
 \email{rdavis@stat.columbia.edu}
 
 \author{Gennady Samorodnitsky}
 \address{School of Operations Research and Information Engineering\\
 	Cornell University}
 \email{gs18@cornell.edu}

 	\thanks{This research was partially
 	supported by  NSF grants DMS-2015379 (Avella Medina and Davis) at Columbia and DMS-2015242 (Samorodnitsky) at Cornell. Part of Samorodnitsky’s research was conducted while visiting Department of Mathematics of National University of Singapore. The hospitality of the department and the university is gratefully acknowledged.}


\begin{document}

\maketitle

\begin{abstract}
We propose kernel PCA as a method for analyzing the dependence structure of multivariate extremes and demonstrate that it can be a powerful tool for clustering and dimension reduction.  Our work provides some theoretical insight into the preimages obtained by kernel PCA, demonstrating that under certain conditions  they can effectively identify clusters in the data.  We build on these new insights to characterize rigorously the performance of kernel PCA based on an extremal sample, i.e., the angular part of random vectors for which the radius exceeds a large threshold.  More specifically, we focus on the asymptotic  dependence of multivariate extremes characterized by the angular or spectral measure  in extreme value theory and provide a careful analysis in the case where the extremes are generated from a linear factor model. We give theoretical guarantees on the performance of kernel PCA preimages of such extremes by leveraging their asymptotic distribution together with Davis-Kahan perturbation bounds. Our theoretical findings are complemented with numerical experiments illustrating the finite sample performance of our methods.

\end{abstract}

\section{Introduction} \label {sec:intro}

 The study and modeling of the behavior of extremes continues to generate increasing interest from scientists in a variety of fields including environmental, industrial, economic and social media related activities. While extremes are reasonably well understood for univariate and low dimensional data, it remains very challenging to model multivariate extremes  when one or more of rare extreme events may occur simultaneously. An important recent line of work in multivariate extreme value theory seeks to connect this literature to ideas from modern statistics and machine learning. This task is not at all trivial since the dependence structure between extreme observations can be very complex and involve notions of dependence that differ from the typical ones arising in the non-extreme world. Work in this direction has included adapting various notions of sparsity for extremes \citep{goixetal2017,meyerandwintenberger2019,simpsonetal2020}, concentration inequalities \citep{goixetal2015,clemencconetal2021}, conditional independence \citep{engelkeandhitz2020, engelkeetal2022}, causality \citep{gneccoetal2021,deuberetal2021} and unsupervised learning \citep{chautru2015,cooleyandthibaud2019,janssenandwan2020,dreesandsabourin2021,avellamedinaetal2021, jalalzaiandleluc2021,fomichovandivanovs2022, rohrbeckandcooley2022}, to name a few important examples. See also \cite{engelkeandivanovs2021} for a review of recent developments in the literature of multivariate extremes. 
 Our work is aligned with this direction of research  as we propose kernel PCA  as a preprocessing tool that facilitates clustering multivariate extremes.

The covariance matrix plays a central role in non-extremal statistics as it is widely used to quantify the linear dependence   among random variables. The eigen-decomposition of the covariance matrix is the building block of principal components analysis (PCA), which in turn is one of the most popular dimension reduction techniques in statistics. It  can be used to find low dimensional projections of $p$-dimensional data into the linear subspace spanned by  $k<p$ eigenvectors associated with the $k$ largest eigenvalues of the empirical covariance matrix. This projection corresponds to the best $k$-dimensional projection in the sense of being the one that retains the most variance present in the original data. Kernel PCA is a nonlinear generalization of PCA that first lifts the original data to a space of functions and then produces low dimensional projections in this function space. This representation can help extract nonlinear structures in the data \citep{scholkopfetal1997}, can be used for data denoising \citep{mikaetal1998} and extracting high dimensional features for regression and classification tasks \citep{rosipaletal2001,kimetal2002, braunetal2008}. More extensive references to kernel methods can be found in the books by \cite{smolaandscholkopt1998} and \cite{steinwartandchristmann2008}.

In this work we use kernel PCA as a denoising tool for multivariate extremes. More specifically, we perform kernel PCA on a subset of observations that is viewed as extremes and we reconstruct the preimages of the kernel PCA projections of this extremal subsample. While kernel PCA projections live in a function space, their preimages live in the original space of the data and constitute our main objects of interest. 

In our analysis, we first provide some general insights showing that kernel PCA preimages naturally cluster in finite subsets of points when there are also some clusters in the kernel space. We believe these insights are interesting in their own right and complement existing work on kernel PCA preimages; see \cite{honeineandcedric2011} for an overview of this literature. We remark that our analysis also complements existing theoretical results regarding the convergence of the spectrum of the empirical covariance operator used in the construction of kernel PCA to a population  covariance operator \citep{shawetayloretal2005,zwaldandblanchard2005,blanchardetal2007}. 

We then consider the case of an extremal sample and utilize tools from multivariate extreme value theory for analyzing the clustering properties of kernel PCA preimages.  In particular, we use multivariate regular variation as a modeling tool since it is closely connected to asymptotic characterizations of multivariate extreme value distributions \citep{resnick2007, resnick2008}.
We provide a detailed analysis of the clustering properties of kernel PCA preimages for multivariate extremes generated by the linear factor model recently introduced by \cite{avellamedinaetal2021}. 
We leverage the asymptotic distributions and rates of convergence derived in the latter work in out perturbation analysis of kernel PCA. Since the spectral measure of this model is discrete, the kernel PCA preimages converge to a well separated set of points in this case. We establish rates of convergence of the kernel matrix defining these preimages and show that they depend on the tail index of the extremes as well as  on the smoothness, around the origin, of the kernel function used for kernel PCA. In the process of establishing these convergence results, we obtain an asymptotic characterization of the point process giving rise to the different clusters of extremes in the linear factor model. We believe this is an interesting side product of our analysis that is likely to be useful in other contexts.

The rest of the paper is organized as follows. Section 2 introduces the framework of kernel PCA needed throughout. Section 3 focuses on the clustering properties of kernel PCA preimages by discussing the special cases of noisy discrete signals which correspond to generative models where there exists a true known correct number of clusters in the data. This perspective reveals that the quality of the clustering of power of the kernel PCA preimages can be controlled by perturbation bounds obtained from a matrix version of the Davis-Kahan theorem. Section 4 focuses the analysis of kernel PCA in the context of multivariate extreme data generated from a heavy-tailed linear factor model. This stylized model is particularly insightful as it asymptotically leads to a discrete angular measure and hence a known finite number of clusters, while in finite samples the extremes are only approximately clustered.  Section 5 discusses the computation of the kernel PCA preimages and Section 6 completes this work with an extensive simulation study that demonstrates that kernel PCA can be a powerful tool for clustering and dimension reduction in a wide variety of models for extremes that not only includes the linear factor model of Section 4, but also other models  where the asymptotic angular measure is continuous.

\section{Background on Kernel PCA}

Kernel principal component analysis builds on the framework of Reproducing Kernel Hilbert Spaces (RKHS). We will review some basic notation and concepts that will be needed in order to describe how kernel PCA gets a low dimensional representation of the data after being lifted to a RKHS.

\begin{definition}
Let $\cH$ be a Hilbert space of real-valued functions defined on a space $\cX$ with  inner product and norm denoted by $\langle\cdot,\cdot\rangle$ and $\|\cdot\|$ respectively. A function   $\kappa:\cX\times\cX\to\mathbb{R}$ is called reproducing kernel if $(i)$ for all $x\in\cX$,  $\kappa(\cdot,x)\in\cH $ and $(ii)$ for all $f\in\cH$, we have $f(x)=\langle f,\kappa(\cdot,x)\rangle$ for all $x\in\cX$. 

If $\cH$ admits a reproducing kernel, then it is called a reproducing kernel Hilbert space.
\end{definition}

By the Moore-Aronszajn Theorem, each symmetric, non-negative definite kernel function $\kappa(\cdot,\cdot)$ on $\cX\times\cX$ can be identified uniquely with a  RKHS of real-valued functions on $\cX$ for which it is the reproducing kernel. For simplicity, in what follows let us denote this RKHS by $\cH$.

The map $\phi:x\mapsto \kappa(\cdot,x)$ from  $\cX$ to $\cH$ is commonly called the  feature map. It follows from the  reproducing kernel property that  $f(x)=\langle f,\phi(x)\rangle$ for all $f\in\cH$ and 
\begin{equation*}
    \langle \phi(x),\phi(y)\rangle=\langle \kappa(\cdot,x),\kappa(\cdot,y)\rangle=\kappa(x,y),\quad \forall x,y,\in\cX.
\end{equation*}
We are now well equipped to describe the methodology of kernel PCA. Assume that we have some observed data $\bx_1,\dots,\bx_n\in\cX\subset \mathbb{R}^d$ and define the empirical covariance operator $\mathscr C_n:\cH\mapsto \cH$ as
\begin{equation*}
\label{eq:cov_operator}
    \mathscr C_nf=\frac{1}{n}\sum_{i=1}^n\phi(\bx_i)\langle \phi(\bx_i),f\rangle=\frac{1}{n}\sum_{i=1}^nf(\bx_i)\phi(\bx_i) .
\end{equation*}
 The covariance operator\footnote{We note that the covariance operator is sometimes defined using the centered feature map $\bar\phi(\bx_i)=\phi(\bx_i)-\frac{1}{n}\sum_{j=1}^n\phi(\bx_j)$ instead of the non-centered feature map $\phi(\bx_i)$. This leads to minor changes to the expressions that we give for the eigenfunctions and eigenvalues. } is positive definite and admits the spectral representation %
\begin{equation*}
    \mathscr C_nf=\sum_{i=1}^n\lambda_i\langle\varphi_i,f\rangle\varphi_i,
\end{equation*}
where $\lambda_1\geq \lambda_2\geq \dots\geq \lambda_n\geq 0$ are the eigenvalues of $\mathscr C_n$ and $\varphi_1,\dots,\varphi_n\in\cH$ are their corresponding normalized eigenfunctions. We are interested in finding the principal components i.e., a lower dimensional representation of arbitrary functions in $\cH$ using the first $m<n$ eigenfunctions of $\mathscr C_n$. More specifically, given a function $f\in\cH$, we want to find its projection into the $m$-dimensional subspace spanned by $\{\varphi_i\}_{i=1}^m$ i.e.
\begin{equation*}
    \cP_mf=\sum_{j=1}^m\lambda_j\langle \varphi_j,f\rangle \varphi_j.
\end{equation*}
The above projection can be computed efficiently  in practice as one can avoid working directly in $\cH$ by noting that the eigenvalues of $\mathscr C_n$ coincide with the eigenvalues of the kernel matrix $C_n=\frac{1}{n}\{  \kappa(\bx_i,\bx_j)\}_{i,j=1}^n$.  
If $\bv_j=(v_{j1},\dots,v_{jn})^\top$ denotes the $j$th eigenvector of $C_n$,  then the $j$th (unnormalized)  eigenfunction of $\mathscr C_n$ can be computed by 
\begin{equation}
\label{eq:eigenfunction}
    \varphi_j=\sum_{i=1}^nv_{ji}\phi(\bx_i).
\end{equation}
From this last formula and the reproducing kernel property we obtain 
\begin{equation*}
   \cP_mf=\sum_{i=1}^m\sum_{j=1}^n\sum_{k=1}^nf(\bx_k)v_{ij}v_{ik}\phi(\bx_j).
   \end{equation*}
%
%
In what follows we focus on the properties of the preimages of the lower dimensional projection of the feature map i.e., $\cP_m\phi$. Intuitively, we would like to understand the impact of the steps: a) lifting the  data from the original space $\cX$ to a function space b) choosing a lower dimensional representation of the lifted data in this richer space, and c) identifying the map of those projections back to the original data space $\cX$.  

We finish this section by mentioning a connection between RKHSs and Gaussian random fields that often provides a convenient and appealing framework to view RKHS and kernel PCA.  Specifically, if $\kappa$ is a stationary symmetric non-negative definite kernel on $\mathbb{R}^d$ (that is, $\kappa(\bx,\by)$
depends only on $\bx-\by$), and $\bm{G}=\bigl( G(\bx), \, \bx\in \bbr^d\bigr)$ is a centered stationary Gaussian
random field, with covariance function $\kappa$, then the RKHS corresponding to $\kappa$ can be identified with the closure in $L^2$ of the space of finite linear combinations of the values of the field at different points (here each $\kappa(\cdot,\bx)$ is identified with $G(\bx)$). This RKHS can also be identified with the subspace of $L^2(\mu)$ consisting of functions with even real parts and odd imaginary parts. Here $\mu$
is the spectral measure of the random field $\bm{G}$, i.e., a finite
symmetric measure on $\bbr^d$ such that
\begin{equation*} 
  \kappa(\bx) = \int_{\bbr^d} e^{i\langle \bx,\bz\rangle}\, \mu(d\bz), \, \bx\in\bbr^d
\end{equation*}
(since $\kappa$ is stationary, we are using a one-argument notation). Here 
$\kappa(\cdot,\bx)$ is identified with $e^{i\langle \cdot, \bx\rangle}$. See e.g., \cite{vanzantenandvandervaart2008}.

 \section{Some insights into kernel PCA preimages}
 
 Let $\calS$ be a subset of  $\bbr^d$, possibly the unit
sphere in $\bbr^d$. Let $S_0\subset \mathcal S$  be a small
subset of $\calS$, potentially of a smaller dimension. Let $T=\{\bt_1,\ldots,
\bt_{n_1}, \bt_{n_1+1}, \ldots, \bt_{n_1+n_2}\}$ be a collection of points
in $\calS$, such that $\bt_1,\ldots, \bt_{n_1}$ lie in or near $S_0$, while
$\bt_{n_1+1}, \ldots, \bt_{n_1+n_2}$ lie at some distance from
$S_0$. Assume, further, that the points $\bt_{n_1+1}, \ldots,
\bt_{n_1+n_2}$ are dispersed, and that $n_2$ is not too large in
comparison with $n_1$. In the following we will focus on RKHS defined by stationary kernels of the form $\kappa(\bx,\by)=R(\bx-\by)$ for all $\bx,\by\in\bbr^d$ and some continuous non-negative definite $R:\mathbb{R}^d\mapsto \mathbb{R}$ (which may be thought of as the covariance function of a stationary Gaussian random field).   





\subsection{Discrete signal case}
 
 Let us first consider a special case. Suppose that $n_2=0$ and that
$S_0=\{ \bs_1,\ldots, \bs_K\}$ is a finite collection of points in
$\calS$. Furthermore, suppose that the points $\bs_1,\ldots, \bs_K$ are well
separated, in the sense that $R(0)=1$ while 
$|R(\bs_i-\bs_j)|$ is small if $i\not=j$ . Let
us suppose, further, that $m_1$ of the points $\bt_1,\ldots, \bt_{n_1}$
 equal $\bs_1$, $m_2$ of the points $\bt_1,\ldots, \bt_{n_1}$
 equal $\bs_2$,  etc. In particular, $n_1=m_1+m_2+\cdots + m_K$.
We will assume for simplicity that the first $m_1$points $\bt_1,\ldots, \bt_{m_1}$
 equal to $\bs_1$, the next $m_2$ points are equal to $\bs_2$, etc.  
In this case the matrix $C_n=\frac{1}{n}\{  R(\bs_i-\bs_j)\}_{i,j=1}^n$  becomes a block
matrix with $K^2$ blocks. The block $(k_1,k_2)$, $k_1,k_2=1,\ldots, K$
has $m_{k_1}$ rows and $m_{k_2}$ columns and $m_{k_1} m_{k_2}$
identical entries equal to
\begin{equation} \label{e:k1k2}
\frac{1}{(m_1+\cdots+m_K)}  R(\bs_{k_1}-\bs_{k_2}).
\end{equation}
In particular, an eigenvector $\bv_j$ of $C_n$
corresponding to any eigenvalue $\lambda_j$ will have the form
\begin{equation} \label{e:solution.blocks}
\bv_j = \bigl( b_1,\ldots, b_1,b_2,\ldots, b_2,\ldots, b_K,\ldots,
b_K\bigr)^\top,
\end{equation}
with each $b_i$ repeated $m_i$ times. 

\bigskip

Returning to the general case, let $\lambda_1\geq \cdots  \geq
\lambda_m \geq 0$ be the $m$ 
largest eigenvalues of $\calC_n$ and let $\varphi_1,\ldots,\varphi_m$ be
the corresponding eigenfunctions. 
Suppose that we now get a new point $\bw\in \mathcal S$. We define its kernel PCA preimage as 
\begin{equation} \label{e:project.back}
  T(\bw) = \text{argmin}_{\bv\in \calS} \| \phi(\bv)-\cP_m\phi(\bw)\|. 
\end{equation}
Note that $\|\phi(\bv)\|^2=R(0)$ is independent of $\bv$. Therefore,
\eqref{e:project.back} reduces to
\begin{align} \label{e:project.alt}
 \notag T(\bw) =& \text{argmax}_{\bv\in \calS} \langle \phi(\bv), \cP_m\phi(\bw)\rangle
    \\
   =& \text{argmax}_{\bv\in \calS}  \sum_{k=1}^m \sum_{\bt_j\in T}
             v_{kj} R(\bw-\bt_j)   \sum_{\bt_j\in T}
             v_{kj} R(\bv-\bt_j). 
\end{align}

\bigskip

\begin{example} \label{ex:m1}
To get a feeling of what is happening let us consider the case
$m=1$.  In this case the problem \eqref{e:project.alt} becomes
\begin{align*} 
 T(\bw) = \mathrm{argmax}_{\bv\in \calS}   \sum_{\bt_j\in T}
             v_{1j} R(\bw-\bt_j)   \sum_{\bt_j\in T}
             v_{1j} R(\bv-\bt_j).
\end{align*}
This means that
\begin{align} \label{e:m1}
  T(\bw) =& \left\{ \begin{array}{ll}
               \mathrm{argmax}_{\bv\in \calS} \sum_{\bt_j\in T}
             v_{1j} R(\bv-\bt_j),
                   & \text{if} \ \sum_{\bt_j\in T}
             v_{1j} R(\bw-\bt_j) >0,\\
         \mathrm{argmin}_{\bv\in \calS}  \,    \sum_{\bt_j\in T}
             v_{1j} R(\bv-\bt_j),
                   & \text{if} \ \sum_{\bt_j\in T}
             v_{1j} R(\bw-\bt_j) <0.   
                 \end{array}
               \right.        
\end{align}               
That is, most of the points $\bw$ get mapped to one of the two points in
$S_0$ that achieve the minimum and maximum \eqref{e:m1}. 
If we return to the special case considered in \eqref{e:k1k2} and
\eqref{e:solution.blocks}, then
\begin{equation*} \label{e:m1.spec}
  \sum_{\bt_j\in T}  v_{1j} R(\bv-\bt_j)  =
 \sum_{k=1}^K m_k b_k R(\bv-\bs_k).
\end{equation*}
Since we are assuming that the points $\bs_1,\ldots, \bs_K$ are well
separated, it is likely that the maximum of that expression will be
achieved near the point $\bs_k$ with the largest value of $m_kb_k$, and
the minumum will be
achieved near the point $\bs_k$ with the smallest value of
$m_kb_k$. That is, most of the points $\bw$ are likely to get mapped
close to one of these two points in the set $S_0$.

\end{example} 

\subsection{Discrete signal with noise}
\label{sec:signal+noise}

Now we consider the generic case where $S_0=\{ \bs_1,\ldots, \bs_K\}$ is  still a finite collection of well separated points  in $\calS$. However, now $n_2>0$ and  the points $\bt_{n_1+1},
\ldots, \bt_{n_1+n_2}$ lie at some distance from $S_0$, and do not
concentrate too much themselves. Now the points
$\bt_1,\ldots, \bt_{n_1}$ are not necessarily exactly equal to one of the points in
$S_0$, but are only lie nearby. Specifically, we assume that  $m_1$ of the
points $\bt_1,\ldots, \bt_{n_1}$ 
are near $\bs_1$, $m_2$ of the points $\bt_1,\ldots, \bt_{n_1}$
are near $\bs_2$,  etc. We still have $n_1=m_1+m_2+\cdots + m_K$.
This time the covariance matrix $C_n$ will have 4
distinct parts whose structure we now describe.

Recall that each $\bt_j,\, j=1,\ldots, n_1$ is near one of the points in
$S_0$. We preserve the numbering we used in \eqref{e:k1k2} and  
\eqref{e:solution.blocks}. That is, we write
\begin{equation} \label{e:tj.n1}
\bt_j=\bs_k+\br_j \ \ \text{if} \ \ m_1+\cdots+m_{k-1}<j\leq
m_1+\cdots+m_{k-1}+m_k, 
\end{equation}
$j=1,\ldots, n_1, \ k=1,\ldots, K$, and assume that $\|\br_j\|$ is small
($\br_j$ does not need to lie in $\calS$).
The matrix $C_n$ will have an $n_1\times n_1$ block matrix in the
top left corner with $K^2$ blocks, whose entries are perturbations of
the entries entries described in \eqref{e:k1k2}. Specifically, the
block $(k_1,k_2)$, $k_1,k_2=1,\ldots, K$ in that matrix 
has $m_{k_1}$ rows and $m_{k_2}$ columns, and the entry in the
position $(i,j)$ within that block can be written as
\begin{equation} \label{e:cblock11}
c_{ij} =  \frac{1}{m_1+\cdots+m_K} 
R(\bs_{k_1}-\bs_{k_2})+ \delta_{ij},
\end{equation}
with
\begin{align} \label{e:delta11}
\delta_{ij} =& \frac{1}{n_1+n_2}\Bigl(
R\bigl(\bs_{k_1}-\bs_{k_2}+\br_i-\br_j\bigr) -
  R\bigl(\bs_{k_1}-\bs_{k_2}\bigr)\Bigr) . 
\end{align} 
We will view this matrix $C_n$ as resulting from a perturbation
of the matrix 
$C^{(0)}_n$ of the same size as $C_n$. The matrix $C^{(0)}_n$ has an
$n_1\times n_1$ block matrix in the 
top left corner with $K^2$ blocks, whose entries are described in
\eqref{e:k1k2}. The rest of the entries of the  matrix $C^{(0)}_n$ are
equal to zero. Let
\begin{equation*} \label{e:Delta}
  \Delta=C_n-C^{(0)}_n
\end{equation*}
be the perturbation. Then $\Delta$ has an
$n_1\times n_1$ block matrix in the 
top left corner with $K^2$ blocks, whose entries are $\delta_{ij}$ in
\eqref{e:delta11}. The rest of the entries of the matrix $\Delta$ have the general form 
$$
\frac{1}{n_1+n_2}R(\bt_i-\bt_j).
$$

We start by noticing that the non-zero eigenvalues of the matrix
$C^{(0)}_n$ coincide with the non-zero eigenvalues of the $n_1\times
n_1$-matrix in its top left corner. Furthermore, the corresponding
eigenvectors of $C^{(0)}_n$ result from taking the eigenvectors of the
latter matrix and appending to them $n_2$ zero entries. We note that the $n_1\times
n_1$-matrix in the top left corner of $C^{(0)}_n$
represents the special situation \eqref{e:k1k2} and we have some understanding of why our procedure results in
points being mapped close to the set $S_0$. 

The true matrix is, of course, $C_n$ and not $C^{(0)}_n$. 
Our plan is to use the Davis-Kahan theorem to check that the
eigenvectors of $C_n$ corresponding to its top eigenvalues are not far
from the eigenvectors of $C^{(0)}_n$ corresponding to its top
eigenvalues. If this is the case, then the eigenfunctions
\eqref{eq:eigenfunction} 
corresponding to the top eigenvalues of the matrix $C_n$ will be close
to the eigenfunctions \eqref{eq:eigenfunction}
corresponding to the top eigenvalues of the matrix $C^{(0)}_n$, and so the algorithm 
will still map most of the points to lie close to the set $S_0$.

We will use a version of the Davis-Kahan theorem given in
\cite{yuetal2015}, which says that, if $\lambda_1,\cdots,
\lambda_m$ are the top eigenvalues of $C_n$ and $\lambda_1^{(0)},\cdots,
\lambda_m^{(0)}$ are the top eigenvalues of $C^{(0)}_n$, then the
corresponding orthonormal eigenvectors $ \bv^{(1)}, \ldots, 
\bv^{(m)}$ and $ \bv^{(0,1)}, \ldots, 
\bv^{(0,m)}$ are close in the following sense. Let $V$ and $V^{(0)}$
be $(n_1+n_2)\times m$ matrices with columns $ \bv^{(1)},
\ldots,  \bv^{(m)}$ and $ \bv^{(0,1)}, \ldots,  
\bv^{(0,m)}$, correspondingly. Then there is an orthogonal 
$m\times m$ matrix $O$
such that
\begin{equation} \label{e:davis-kahan}
  \| VO-V^{(0)}\|_{\rm F}\leq \frac{2\min\bigl( m^{1/2} \|
    \Delta\|_{\rm op}, \, \|
    \Delta\|_{\rm F}\bigr)}{\lambda_m^{(0)}-\lambda_{m+1}^{(0)}}. 
\end{equation}  
Here $\| A\|_{\rm F}$ and $\| A\|_{\rm op}$ are, respectively, the
Frobenius norm and the operator norm of a matrix $A$. This is Theorem
2 in \cite{yuetal2015}.

Since the orthogonal matrix $O$ in \eqref{e:davis-kahan} plays no role
in the projection onto the subspace spanned by the eigenfunctions
corresponding to the top eigenvectors, we need to check that the bound
in the right hand side of \eqref{e:davis-kahan} is small. Assuming that the covariance matrix of the
Gaussian random field is nonsingular, in the generic case the matrix $C^{(0)}_n$ will have
    $K$ nonzero eigenvalues. Furthermore, the size of these $K$
    eigenvalues should be 
    comparable to the size of the entries in the  matrix
    $C^{(0)}_n$, which by \eqref{e:k1k2}  are of order $1/n$. It is
    likely that also the distances between these eigenvalues are of
    the same order. This, of course, means that we should choose
    $m\leq K$ and, ideally, $m=K$. In this case we expect the
    denominator in the right hand side of \eqref{e:davis-kahan} to be
    of order $1/n$. 

    Now consider the numerator  in the right hand side of
    \eqref{e:davis-kahan}. One can expect that the norms appearing
    there would be comparable to the size of the entries in the matrix
    $\Delta$. Notice that the entries in this matrix that are not in
    the $n_1\times n_1$ block matrix in the top left corner are still
    of the order $1/n$, but because the points $\{\bs_k\}$ are well
    separated, these entries will  be small in comparison with
    the denominator in the right hand side of
    \eqref{e:davis-kahan}. Finally, the entries in the $n_1\times n_1$
    block matrix in the top left corner of the matrix
    $\Delta$ will be small if the perturbations $\{r_i\}$ are small, and
    this should be established on the case-by-case basis, for different data-producing mechanisms. In the next section we apply this idea to the extremes of a heavy tailed linear factor model.

    \section{Applications to the linear factor model}

Consider the linear factor model
\begin{equation} \label{e:LFM}
\BX=A\BZ,
\end{equation}
where $A$ is a $d\times p$ 
matrix of non-negative elements and $\BZ$ is a $p$-dimensional
random vector of factors consisting of independent and identically
distributed non-negative random variables\footnote{The results of this section can be extend with simple but tedious modifications to symmetric regularly varying iid random variables and a real-valued $A$.}, 
that  have asymptotically Pareto tails, i.e.,
\begin{equation} \label{e:pareto}
  \bbP(Z_1>z)\sim c_\alpha z^{-\alpha},~~\mbox{as $z\to\infty,$}
\end{equation}
for some $\alpha>0$ and $c_\alpha>0$.  
As pointed out in \cite{avellamedinaetal2021}, it follows immediately from \eqref{e:LFM} and \eqref{e:pareto} (see, for example, \cite{basrak2002characterization}, Proposition A.1) that
$\BX$ is a multivariate regularly varying
random vector satisfying  
\begin{equation*} \label{e:multv.regvar}
\lim_{x\to\infty} \bbP\left( \ \frac{\BX}{\|\BX\|}\in\cdot~|~\|\BX\|>x\right) 
\Rightarrow  \Gamma(\cdot)\,,
\end{equation*} 
where $\Rightarrow$ denotes weak convergence on the unit sphere $\bbS^{d-1}$, 
 $\Gamma$ is the discrete probability measure on $\bbS^{d-1}$ given by
\begin{equation*} \label{e:spectral.m}
  \Gamma(\cdot) =w^{-1}\sum_{k=1}^p \|\ba^{(k)}\|^\alpha
       \delta_{\frac{\ba^{(k)}}{\|\ba^{(k)}\|}}(\cdot)\,,
\end{equation*}
where 
$\delta_x(\cdot)$ is the Dirac measure that puts unit mass at $x$, $\ba^{(k)}$ is the $k$th column of the matrix $A$, $k=1,\ldots, p$, and  
\begin{equation} \label{e:total.w}
w=\sum_{k=1}^p \|\ba^{(k)}\|^\alpha\,.
\end{equation}
Based on a random sample of iid copies of $\BX_1,\ldots,
\BX_n$ of $\BX$ as above, we expect for large $n$, the {\it angular parts} $\BX_i/\|\BX_i\|$ of the sample  for which $\|\BX_i\|$ is large, to cluster around the points 
\begin{equation*} \label{e:centers.LFM}
  \bs_k=\frac{\ba^{(k)}}{\|\ba^{(k)}\|}, \ k=1,\ldots, p,
\end{equation*}
as in Section \ref{sec:signal+noise}.  

In order to understand how well the kernel PCA algorithm works for extreme values in this model, we will analyze the corresponding perturbation matrix $\Delta$. We start with the $n_1\times n_1$ block matrix in the top left corner
of $\Delta$, whose entries are given by
\eqref{e:delta11}. We call this matrix $\Delta_B$. 

The first result that we will need in the study of $\Delta_B$ is a characterization of the convergence of the covariance function evaluated at the difference of directions of extreme observations. 
We will assume that 
for some $\theta\in [1,2] $ and $d_\theta>0$, 
\begin{equation} \label{e:cov.zero}
  R(\0)-R(\bx) \sim d_\theta\| \bx\|^\theta \ \ \text{as $\bx\to \0$.}
\end{equation}
Common choices of $R$ are the exponential covariance function $R(\bx)=\exp\bigl\{ -\gamma \|\bx\|\bigr\}$ for which  $\theta=1$, and the Gaussian covariance function $R(\bx)=\exp\bigl\{ -\gamma \|\bx\|^2\bigr\}$, for which $\theta=2$. 


\begin{theorem} \label{pr:Delta.B} Let $(u_n)$ be a sequence of levels such that $u_n\to\infty$ and suppose the covariance function $R$ satisfies \eqref{e:cov.zero} and is continuously differentiable outside of the origin. 
	
{\rm (i)} \ Let $i\not=j$ belong to a diagonal block $(k,k)$ of the matrix
$\Delta_B$, $k=1,\ldots, p$. Then for any $i\not=j$, computing the law in the left hand side as the 
conditional law given the event $\{\|\BX_i\|>u_n,\|\BX_j\|>u_n,\,\, Z_{ik}>u_n/w^{1/\alpha},\, Z_{jk}>u_n/w^{1/\alpha}\}$, ($w$ defined in \eqref{e:total.w}), 
\begin{equation} \label{e:diag}
u_n^\theta\Bigl[ R(\0)- R\bigl(\BX_i/ \|\BX_i\|-\BX_j/
\|\BX_j\|\bigr)\Bigr] \Rightarrow d_\theta \|\bS^{(k)}_1-\bS^{(k)}_2\|^\theta,
\end{equation}
where $\bS^{(k)}_1,\bS^{(k)}_2$ are independent random vectors with the common law
defined as the law of
\begin{equation} \label{e:lim.law.k}
\frac{1}{w_k^2W_\alpha}\bigl( S^*_{1,-k},\ldots,
S^*_{d,-k}\bigr)^T\,,
\end{equation}
with $w_k=\|\ba^{(k)}\|$.  Here, with $\BX$ as in \eqref{e:LFM}, we set  
$$
S^*_{l,-k}=\sum_{r=1}^d \bigl( a_{rk}^2 X_{l,-k} -a_{lk}a_{rk}X_{r,-k}\bigr)
$$
with
$$
X_{l,-k} = X_l-a_{lk}Z_k, \ \
l=1,\ldots, d.
$$
Furthermore, 
$W_\alpha$ is standard Pareto$(\alpha)$ random variable  ($\bbP(W_\alpha>x)=x^{-\alpha}, \,
x\geq 1$), independent of $Z_1,\ldots, Z_p$.

{\rm (ii)} \ Let $(i,j)$ belong to a  block $(k_1,k_2)$ of the matrix
$\Delta_B$, $k_1\not= k_2, \, k_1,k_2=1,\ldots, p$. Then, for any $i\not=j$, computing the law in the left hand side as the 
conditional law given the event
  $\{\|\BX_i\|>u_n,\|\BX_j\|>u_n,\,\, Z_{ik_1}>u_n/w^{1/\alpha},\, Z_{jk_2}>u_n/w^{1/\alpha}\}$,
\begin{equation} \label{e:lim.law.k12}
u_n\Bigl[ R(\bs_{k_1}-\bs_{k_2})- R\bigl(\BX_i/ \|\BX_i\|-\BX_j/
\|\BX_j\|\bigr)\Bigr] \Rightarrow \langle \nabla
R(\bs_{k_1}-\bs_{k_2}),\bS^{(k_2)}_1-\bS^{(k_1)}_2\rangle,
\end{equation}
with $\bS^{(k_1)}_1,\bS^{(k_2)}_2$ independent and distributed as above. 
\end{theorem}
\begin{proof}
(i) \ 
  Write
\begin{align*}
   &R(\0)- R\bigl(\BX_i/ \|\BX_i\|-\BX_j/
\|\BX_j\|\bigr) \\
=& R(\0)-R\Bigl[ u_n^{-1}\Bigl( u_n\bigl( \BX_i/
\|\BX_i\|-\bs_k\bigr) -u_n\bigl( \BX_j/
\|\BX_j\|-\bs_k\bigr)\Bigr)\Bigr]\,.
\end{align*}
By Theorem 4.1 in \cite{avellamedinaetal2021} (which holds under the  sole assumption $u_n\to\infty$), we have
\begin{equation} \label{e:from.spectr}
u_n\bigl( \BX_i/
\|\BX_i\|-\bs_k\bigr) \Rightarrow \bS_1^{(k)} 
\end{equation} 
weakly in $\bbr^d$. The same is true when $i$ is replaced by $j$ and by independence 
and \eqref{e:cov.zero}, this implies 
\eqref{e:diag} using the delta method.

(ii) \ Write
\begin{align*}
&R(\bs_{k_1}-\bs_{k_2})- R\bigl(\BX_i/ \|\BX_i\|-\BX_j/
\|\BX_j\|\bigr) \\
=& R(\bs_{k_1}-\bs_{k_2})- R\Bigl[ \bs_{k_1}-\bs_{k_2} 
 + u_n^{-1}\Bigl( u_n\bigl( \BX_i/
\|\BX_i\|-\bs_{k_1}\bigr) -u_n\bigl( \BX_j/
\|\BX_j\|-\bs_{k_2}\bigr)\Bigr)\Bigr].
\end{align*}
Now \eqref{e:lim.law.k12} follows from the differentiability of $R$
outside of the origin and  \eqref{e:from.spectr}, once again using the delta method.

\end{proof}

\bigskip

Recall that we would like to understand the magnitude of the matrix norms appearing in the numerator of \eqref{e:davis-kahan}. As explained above, we expect the two norms to be of the same order, so we will consider the Frobenius norm of $\Delta$. We start with the matrix $\Delta_B$.  Theorems \ref{thm:FrobScen1}, \ref{thm:FrobScen2} and  \ref{thm:FrobScen.mid} below constitute the main results regarding the asymptotic behaviour of the Frobenius norm of $\Delta_B$ under the linear factor model. These theorems highlight that there are three different regimes resulting from the tail index of the underlying factors and the smoothness of the chosen kernel function. In particular, the regime  $\alpha<2\theta$ requires us to establish a new point process convergence result stated in Theorem \ref{pr:point_process}. It serves as an important technical tool in the proof of Theorems \ref{thm:FrobScen2} and  \ref{thm:FrobScen.mid}, and we believe that it can be of broader interest.

It will be convenient to introduce some notation used in \cite{avellamedinaetal2021}. For $n=1,2,\ldots,$ we define the set of indexes corresponding to extreme observations 
\begin{equation*} \label{e:In}
  \mathcal{I}_n=\bigl\{ i=1,\ldots, n:\, \|\BX_i\|>u_n\bigr\},
\end{equation*}
and denote its cardinality by  $N_n=\text{card}(\mathcal{I}_n)$. Now 
assuming the thresholds $(u_n)$  satisfy the standard assumptions 
\begin{equation} \label{e:level.usual}
	u_n\to\infty, \ \ n^{-1/\alpha}u_n\to 0, \ \ n\to\infty\,,
\end{equation}
which imply $n\PP(\|\BX\|>u_n)\sim c_\alpha (n^{-1/\alpha}u_n)^{-\alpha}\to \infty$,
it follows  from  \eqref{e:pareto} and \eqref{e:level.usual} (see \cite{avellamedinaetal2021}) that the mean and variance of $N_n/(nu_n^{-\alpha})$ converge to $cw$ and $0$, respectively and hence
that
\begin{equation} \label{e:In.size}
N_n/(nu_n^{-\alpha})\cip cw,\mbox{~as $n\to\infty$.}
\end{equation}
Let $\calI^{(k)}_n$ denote the collection of indexes of extremes caused by the $k$th factor
$Z_k$, $k=1,\ldots, p$. Formally,
\begin{equation*} \label{e:clusters}
\calI^{(k)}_n = \bigl\{ i=1,\ldots, n:\, \| \BX_i\|>u_n, \,
Z_{ik}>u_n/w^{1/\alpha}\bigr\}, \ k=1,\ldots, p
\end{equation*}
(this is (4.22) in \cite{avellamedinaetal2021}).
We denote $N_n^{(k)}=\text{card}(\calI^{(k)}_n), \, k=1,\ldots, p$. It turns out that with probability converging to 1, the sets
$\calI^{(1)}_n, \ldots,  \calI^{(p)}_n$ are disjoint and 
$N_n=N_n^{(1)}+ \cdots + N_n^{(p)}$; see 
Lemma 4.4 in \cite{avellamedinaetal2021}. Furthermore, by
(4.24) in \cite{avellamedinaetal2021}, 
\begin{equation} \label{e:N.n}
N_n^{(k)}/(nu_n^{-\alpha}) \cip c_\alpha w_k^\alpha, \
k=1,\ldots, p, \mbox{~as $n\to\infty$,}
\end{equation}
 where we recall 
\begin{equation*} \label{e:w}
  w = \sum_{k=1}^p \|\ba^{(k)}\|^\alpha=\sum_{k=1}^p w_k^\alpha.
\end{equation*}
Using this notation, we see that
\begin{align} \label{e:Frob.B}
\notag   \|\Delta_B\|_{\rm F} =&
      \left(\sum_{i=1}^{n_1}\sum_{j=1}^{n_1}\delta_{ij}^2\right)^{1/2} 
                           = \left(\sum_{k_1=1}^p \sum_{k_2=1}^p
                           \sum_{i\in \calI^{(k_1)}_n}
                           \sum_{j\in \calI^{(k_2)}_n}   \delta_{ij}^2 \right) ^{1/2} \\
  =:& \left(\sum_{k_1=1}^p \sum_{k_2=1}^p
             F_{k_1,k_2}(n) \right)^{1/2}.
\end{align}
Further note that $F_{k_1,k_2}(n)$ can be written as a $U$-statistic of the form 
\begin{align} \label{e:F.k1k2}
\notag F_{k_1,k_2}(n) =& \frac{1}{N_n^2} \sum_{i\in \calI^{(k_1)}_n}
                           \sum_{j\in \calI^{(k_2)}_n}
  \left[R\left(\frac{\BX_i}{ \|\BX_i\|}-\frac{\BX_j}
              {\|\BX_j\|}\right)-R(\bs_{k_1}-\bs_{k_2})\right]^2 \\
\notag =&  \frac{1}{N_n^2} \sum_{i=1}^{N_n^{(k_1)}}
                           \sum_{j=1}^{N_n^{(k_2)}}
   \left[R\left(\BY_i^{(k_1)}-\BY_j^{(k_2)}\right)-R(\bs_{k_1}-\bs_{k_2})\right]^2 \\
 \notag =& \frac{N_n^{(k_1)}N_n^{(k_2)}}{N_n^2} 
\frac{1}{N_n^{(k_1)}N_n^{(k_2)}} \sum_{i=1}^{N_n^{(k_1)}}
                           \sum_{j=1}^{N_n^{(k_2)}}
 \left[R\left(\BY_i^{(k_1)}-\BY_j^{(k_2)}\right)-R(\bs_{k_1}-\bs_{k_2})\right]^2 \\
 =&:  \frac{N_n^{(k_1)}N_n^{(k_2)}}{N_n^2} G_{k_1,k_2}(n).
\end{align}
%
In \eqref{e:F.k1k2}, for each $k=1,\ldots, p$, we enumerate $\BX_{i}/\|\BX_{i}\|, 
\, i\in \mathcal{I}_n^{(k)}$ as $\BY_i^{(k)}, \, i=1,\ldots, N_n^{(k)}$, a sample on $\bbS^{d-1}$
of random size $N_n^{(k)}$. We also have that
\begin{equation} \label{e:ratio.N12}
 \frac{N_n^{(k_1)}N_n^{(k_2)}}{N_n^2}   \cip \frac{w_{k_1}^\alpha w_{k_2}^\alpha
  }{w^2} \mbox{~as $n\to\infty$.}
\end{equation}
Theorem \ref{pr:Delta.B} makes it reasonable to expect that under
some assumptions it should be true that
\begin{equation} \label{e:kk.scen1}
u_n^{2\theta}  G_{k,k}(n) \to d_\theta^2
\E\bigl\|\bS_1^{(k)}-\bS_2^{(k)}\bigr\|^{2\theta}, 
\end{equation}
and that for $k_1\not= k_2$,
\begin{equation} \label{e:k12.scen1}
u_n^{2}  G_{k_1,k_2}(n) \to 
\E\bigl[\langle \nabla
R(\bs_{k_1}-\bs_{k_2}),\bS^{(k_2)}_1-\bS^{(k_1)}_2\rangle\bigr]^2,
\end{equation}
at least in probability.  At the very least \eqref{e:kk.scen1} requires
$\E\bigl\|\bS_1^{(k)}\bigr\|^{2\theta}<\infty$, while \eqref{e:k12.scen1} requires
$\E\bigl\|\bS_1^{(k)}\bigr\|^{2}<\infty$. Since $\theta\geq 1$, we will assume that 
\begin{equation} \label{e:alpha.theta.1}
  \alpha>2\theta.
\end{equation}
The following statement formalizes this intuition and characterizes the behavior of $\|\Delta_B\|_{\rm F}$ when the tails are
not too heavy i.e., when \eqref{e:alpha.theta.1} holds. The proofs of this and the subsequent theorems in this section are given in the Appendix .

\begin{theorem} \label{thm:FrobScen1}
  Suppose that \eqref{e:alpha.theta.1} holds. Then,
  \begin{equation}
      \begin{split} \label{e:FrobScen1}
 u_n^2  \|\Delta_B\|_{\rm F} \cip & \,\frac1w\biggl( \sum_{k=1}^p \bigl(
   d_*  w_k^\alpha)^2   \mathbb{E}\bigl\|\bS_1^{(k)}-\bS_2^{(k)}\bigr\|^{2}   \\
 & 
   \quad\quad+ \underset{k_1\not= k_2}{\sum_{k_1=1}^p\sum_{k_2=1}^p} w_{k_1}^\alpha w_{k_2}^\alpha
   \E\bigl[\langle \nabla
R(\bs_{k_1}-\bs_{k_2}),\bS^{(k_2)}_1-\bS^{(k_1)}_2\rangle\bigr]^2
 \biggr)^{1/2}, \mbox{as $n\to\infty$},
  \end{split}
  \end{equation}
 where $d_*=d_\theta$ if $\theta=1$ and $d_*=0$ if
  $\theta>1$.   
  
\end{theorem}

Note that in case $\theta>1$ (impliying that the covariance function $R(\cdot)$ is differentiable at the origin) only the off-diagonal terms contribute to the asymptotic behavior of the Frobenius norm.  
In the scenario 
\begin{equation} \label{e:alpha.theta.2}
  \alpha<2\theta\,,
\end{equation}
the analysis for the diagonal and off-diagonal terms is different. We will show that under the following additional assumption on the sequence of levels, 
\begin{equation} \label{e:level.unusual}
  n^{-1/\alpha}u_n^2\to\infty, \ \ n\to\infty.
\end{equation}
upon proper rescaling, these terms converge in distribution to $\alpha/(2\theta)$-stable positive random variables.  

We start with a result on convergence of a certain sequence of point processes that may be of independent interest. For every 
 $k=1,\ldots, p$ and $n\geq 1$ we define a point process  
\begin{equation*} \label{e:pprocess.n}
  M_n^{(k)}=\sum_{i=1}^{N_n^{(k)}} \delta_{u_n^2n^{-1/\alpha}(\BY^{(k)}_i-\bs_k)}.
\end{equation*}

 \begin{theorem}\label{pr:point_process}
 Suppose that \eqref{e:alpha.theta.2} holds, and that the sequence of levels satisfies \eqref{e:level.usual} and 
 \eqref{e:level.unusual}. Then,
 \begin{equation} \label{e:pp.conv}
  M_n^{(k)}\Rightarrow M_\alpha^{(k)}, \ \ n\to\infty,
\end{equation}
weakly in the vague topology on $\bbr^d\setminus\{0\}$, where
$M_\alpha^{(k)}$ is a Poisson point process on $\bbr^d$ with  mean
measure
\begin{equation*} \label{e:mean.m}
  m_\alpha^{(k)}(\cdot) = (c_\alpha^2 w_k^{2}/2) \sum_{j\not= k} m_{\alpha,j}(\cdot),
\end{equation*}
and for $j=1,\ldots, p$, $j\not= k$,
\begin{equation*} \label{e:mean.ray}
m_{\alpha,j} = \int_0^\infty \alpha y^{-(1+\alpha)}
\delta_{y\boldb^{(j,k)}}(\cdot) dy.
\end{equation*}
Here $\boldb^{(j,k)}=\bigl( b^{(j,k)}_1,\ldots, b^{(j,k)}_d\bigr)^\top$,
and for $r=1,\ldots, d$, 
\begin{equation*} \label{e:bb}
b^{(j,k)}_r =w_k\left( a_{rj}-\frac{a_{rk}}{w_k^2} \sum_{m=1}^d
  a_{mj}a_{mk}\right). 
\end{equation*}
Moreover, the convergence in \eqref{e:pp.conv} is joint in $k=1,\ldots,p$, i.e.,
\begin{equation*} \label{e:pp.conv2}
  (M_n^{(1)},\ldots,M_n^{(p)})\Rightarrow (M_\alpha^{(1)},\ldots,M_\alpha^{(p)})\,, \ \ n\to\infty,
\end{equation*}
where $M_\alpha^{(1)},\ldots,M_\alpha^{(p)}$ are independent Poisson point processes.
\end{theorem}


The asymptotic behaviour of the Frobenius norm of the perturbation matrix $\Delta_B$ can be now expressed via integrals with respect to the limit point processes $M_\alpha^{(k)}$. We consider two separate cases. 

\begin{theorem}\label{thm:FrobScen2}
  Suppose that $\alpha<2$ and that the sequence of levels satisfies \eqref{e:level.usual} and 
 \eqref{e:level.unusual}. Then, 
  \begin{align*} \label{e:FrobScen2}
 \bigl( u_n^{2-\alpha/2}n^{-1/\alpha+1/2}\bigr) &\|\Delta_B\|_{\rm F}
   \Rightarrow \Biggl( \frac{2d_*^2}{c_\alpha}\sum_{k=1}^p  w_k^\alpha\int_{\bbr^d} 
  \|x\|^{2\theta} M_\alpha^{(k)}(d\bx)   \\
 \notag & 
          + \frac{1}{w^2c_\alpha}\underset{k_1\not= k_2}{\sum_{k_1=1}^p\sum_{k_2=1}^p}
 \int_{\bbr^d} \Bigl[\bigl( \nabla
R(s_{k_1}-s_{k_2}),  x\bigr)\Bigr]^2 \,
                     \bigl( w_{k_2}^\alpha M_\alpha^{(k_1)}+  w_{k_1}^\alpha M_\alpha^{(k_2)}\bigr)
(d\bx)  \Biggr)^{1/2}, 
  \end{align*}
  where $d_*=d_\theta$ if $\theta=1$ and $d_*=0$ if
  $\theta>1$. 
\end{theorem}
Once again the off-diagonal terms dominate for the case $\theta>1$.

In the final situation we consider in this section we have, once again, convergence in probability. 

\begin{theorem} \label{thm:FrobScen.mid}
 Suppose that $2<\alpha<2\theta$ and that the sequence of levels satisfies \eqref{e:level.usual} and 
 \eqref{e:level.unusual}.       
 Then
  \begin{equation*}
      \label{e:FrobScen.mid}
 u_n  \|\Delta_B\|_{\rm F} \to \frac1w\Biggl(  
    \underset{k_1\not= k_2}{\sum_{k_1=1}^p\sum_{k_2=1}^p} {k_1}^\alpha
  w_{k_2}^\alpha \E\bigl[\bigl( \nabla
R(s_{k_1}-s_{k_2}),S^{(k_1)}_1-S^{(k_2)}_1\bigr)\bigr]^2
 \Biggr)^{1/2}
  \end{equation*} 
  in probability. 
\end{theorem}

The theoretical results of this section rigorously establish the precise rate at which the Frobenius norm of the perturbation matrix vanishes when the sample size tends to infinity, when performing kernel PCA on data generated from a linear factor model. This explains why kernel PCA preimages correctly find the underlying clusters of extremes in this model. Our numerical experiments will show that empirically kernel PCA performs well in many additional settings, including models where the angular measure is continuous. Our experiments rely on the gradient-based optimization framework discussed next.

\section{Computational considerations}

Our task is to obtain low-dimensional kernel PCA  representations of the observations in their natural domain (the unit sphere in the case of extremes), usually referred to as kernel PCA preimages (of the low-dimensional representation of the observations in the RKHS). 
There have been numerous proposals for recovering such kernel PCA preimages, including a fixed point iteration in \cite{mikaetal1998}, the multidimensional scaling-based procedure of  \cite{kwokandtsang2004}, and  penalized methods of  \cite{bakiretal2004, zhengetal2010}, among others; see \cite{honeineandcedric2011} for an overview.

We formally define a preimage as a solution to the optimization problem   \eqref{e:project.alt}, which we repeat for convenience here:
\begin{align} \label{eq:preimage1} 
  T(\bw) = \text{argmax}_{\bv\in \calS}  \sum_{k=1}^m \sum_{\bt_j\in T}
             v_{kj} R(\bw-\bt_j)   \sum_{\bt_j\in T}
             v_{kj} R(\bv-\bt_j). 
\end{align}
One can in principle solve the problem above by Monte Carlo up to arbitrary numerical precision. However, since this involves  maximization over a potentially high-dimensional sphere is involved, computational issues are important. 
    
%

In our implementation we employed projected gradient descent, which is a standard algorithm for optimizing  smooth objective functions under convex constraints. Denoting the function being maximized in   \eqref{eq:preimage1} by $f(\bv)$, the algorithm is defined by the iterates 
$$ \bv^{(k)}=\Pi_{\mathcal S}(\bv^{(k-1)}-\eta \nabla f(\bv^{(k-1)})),$$
where $\eta>0$ is a fixed step-size parameter, $\nabla$ is the gradient and the projection operator $\Pi$ to $\mathcal{S}$ is defined as 
$$\Pi_{\mathcal S}(\bx)=\underset{\by\in \mathcal S}{\mbox{argmin}}\,\|\bx-\by\|_2. $$
%


In the sequel we use the Gaussian kernel $R(\bx)=\exp(-\gamma\|\bx\|^2_2)$, as it is perhaps the most popular kernel function used in machine learning.  
 
%
 In order to implement this algorithm, one also needs to calculate the Lipschitz constant $\beta$ of the function $f(\bv)$ in order to set a stepsize $\eta\leq 1/\beta$. 
A direct calculation shows that, in the case of the Gaussian kernel, it suffices to set the stepsize to be 
$$
\eta= \left( 2 \left\|\sum_{k=1}^m \sum_{\bt_j\in T}
             \bv_jv_{kj} R(\bw-\bt_j)     \right\| \right)^{-1}
$$ 
in order to ensure the converge of projected gradient descent. In the last equation $\bv_j=(v_{j1},\dots,v_{jd})^\top$ denotes the $j$th
eigenvector of the kernel matrix $C_n=\{R(\bx_i-\bx_j)\}_{i,j=1}^n$.

\section{Empirical study}
 
We have chosen several numerical examples to illustrate the performance of kernel PCA in scenarios that go well beyond the linear  factor model studied in detail in Section 4. In particular, we consider a contaminated linear factor model and three examples where the spectral measure is continuous.  
In all the examples considered below we compute weighted adjacency matrices using the Gaussian kernel $\kappa(\bx,\by)=\exp(-\gamma\|\bx-\by\|^2)$ with $\gamma=1$ unless explicitly stated otherwise. We select the number $m$ of the largest eigenvalues of the covariance operator to use kernel PCA 
as suggested by the screeplots of the kernel matrix. In most of the examples with a well-defined true number of clusters of extremes, the choice of $m$ suggested by the screeplot matched that number. In all our examples we generate $10,000$ observations and take a sample of extremes defined as the $200$ observations with the largest Euclidean norms.   
 
 In all the examples below the random vector of interest $\bX$ is regularly varying, a commmon assumption in studying  heavy-tailed data.  This means $\|\bX\|$ is regularly varying with index $\alpha>0$  and the angular part $\bX/\|\bX\|$ is independent of the radius as the radius becomes large. Formally,
\begin{eqnarray*}\label{eq:mrv}
\lim_{r\to\infty} \bbP\bigl( \BX/\|\BX\|\in \cdot\,\mid \, \|\BX\|>r\bigr)\Rightarrow \Gamma(\cdot)
\end{eqnarray*}
and
\begin{eqnarray*}\label{eq:mrv2}
\lim_{r\to\infty}\frac{\bbP\bigl( \|\BX\|>rx\bigr)}{\bbP\bigl( \|\BX\|>r\bigr)}
=x^{-\alpha}
\end{eqnarray*}
for all $x>0$. The limit probability measure $\Gamma$ is called the {\it angular measure} or {\it spectral measure} and describes how likely the
extremal observations are to point in different
directions. In other words, the angular measure describes the limiting extremal angle for high threshold exceedances that correspond to large $\|\bX\|$.  The support of this measure is particularly
important since it shows which directions of the extremes are feasible
and which are not feasible.

\subsection{Contaminated linear factor model}
 
 The extremes from the linear factor model 
 \eqref{e:LFM} have a discrete spectral measure and we expect kernel PCA applied to these extremes to concentrate the extremes near the atoms of the spectral measure with the largest masses. What happens if we``contaminate'' this spectral measure by a small continuous component? We investigate this question empirically by considering the extremes arising from the model 
 \begin{equation*}
     \bX_i= A \bZ_i+\sigma\bvarepsilon_i, \quad i=1,\dots,n,
 \end{equation*}
 with $\{\BZ_i\}_{i=1}^n$ i.i.d. copies of the vector $\BZ$ in \eqref{e:LFM}, and $\bm{\varepsilon}$ is the ``contamination'' vector. In this case $\sigma\geq0$ regulates the level of contamination. We choose the vector $\BZ$ to consist of i.i.d. standard Fr{\'e}chet\footnote{The standard Fr{\'e}chet distribution is $F(x)=e^{-x^{-1}},~x\ge 0$.} components  and obtain  $\bm{\varepsilon}$ by multiplying the element-wise absolute values of a standard $p$-dimensional normal random vector by 
 univariate standard Fr{\'e}chet random variable (with all random objects independent). The contamination adds a uniform component to the spectral measure, the weight of which is proportional to $\sigma$. 
   In our simulation we take $d=4,\, p=2$, and use the matrix
   \begin{equation}\label{eq:A}
    A= \begin{pmatrix} 
 0.1 & 0.9\\
 0.2 & 0.8\\
 0.3 & 0.7 \\
 0.4 & 0.6 \\
 \end{pmatrix}
 \end{equation}
 and choose $\sigma=1$, which leads to a sample of extremes where approximately half can be assigned to the signal of the latent factors.
 
\begin{figure}[h!]
    \centering
    \includegraphics[scale=0.35]{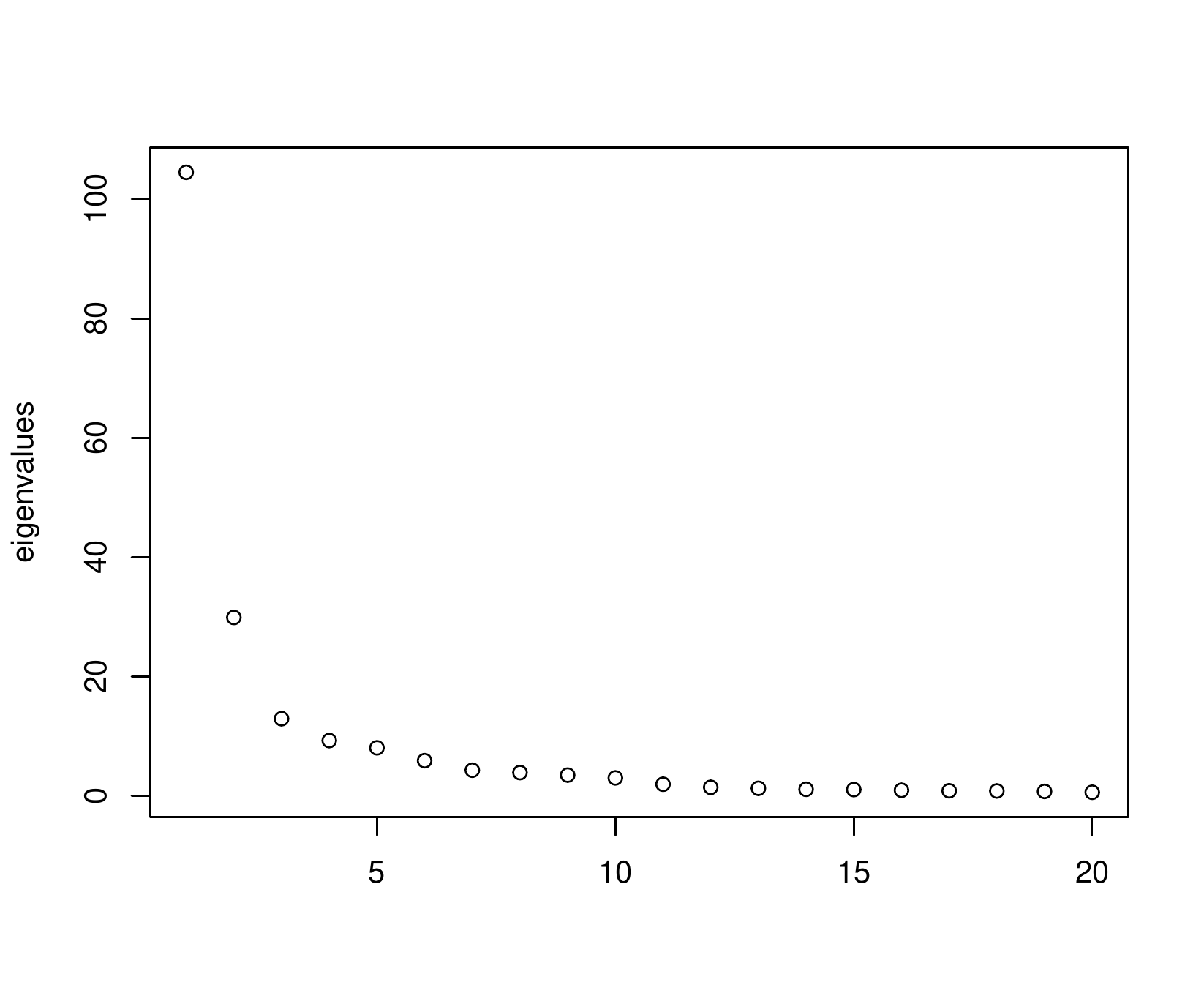}
    \caption{{\small Largest 20 eigenvalues of the kernel matrix used to run kernel PCA on extremes from a contaminated linear factor model}}
    \label{fig:max-linear_screeplot}
\end{figure}

\begin{figure}[h!]
    \centering
    \hfill
\subfigure[Contaminated linear factor model data]{\includegraphics[scale=0.42]{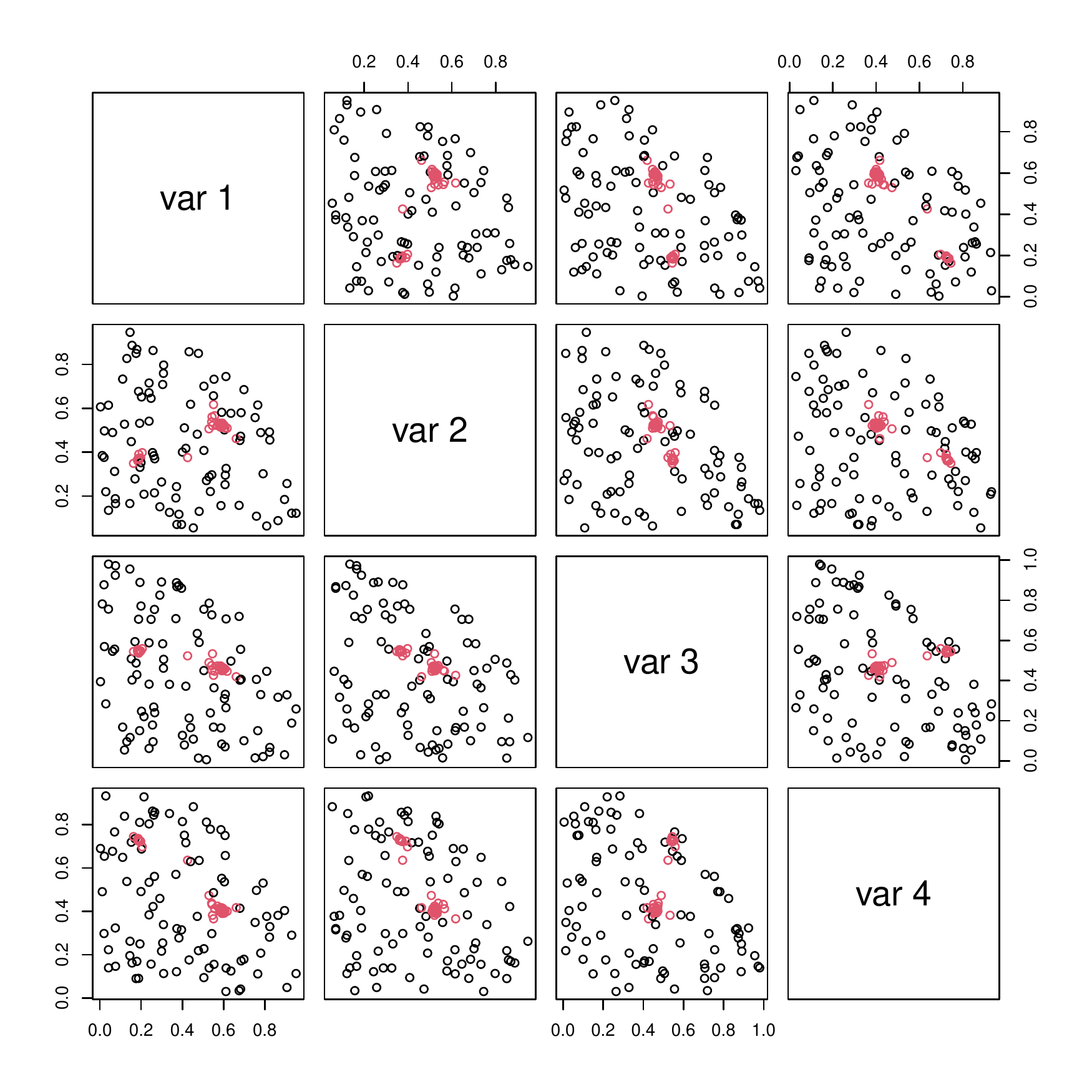}}
\hfill
\subfigure[Preimages]{\includegraphics[scale=0.42]{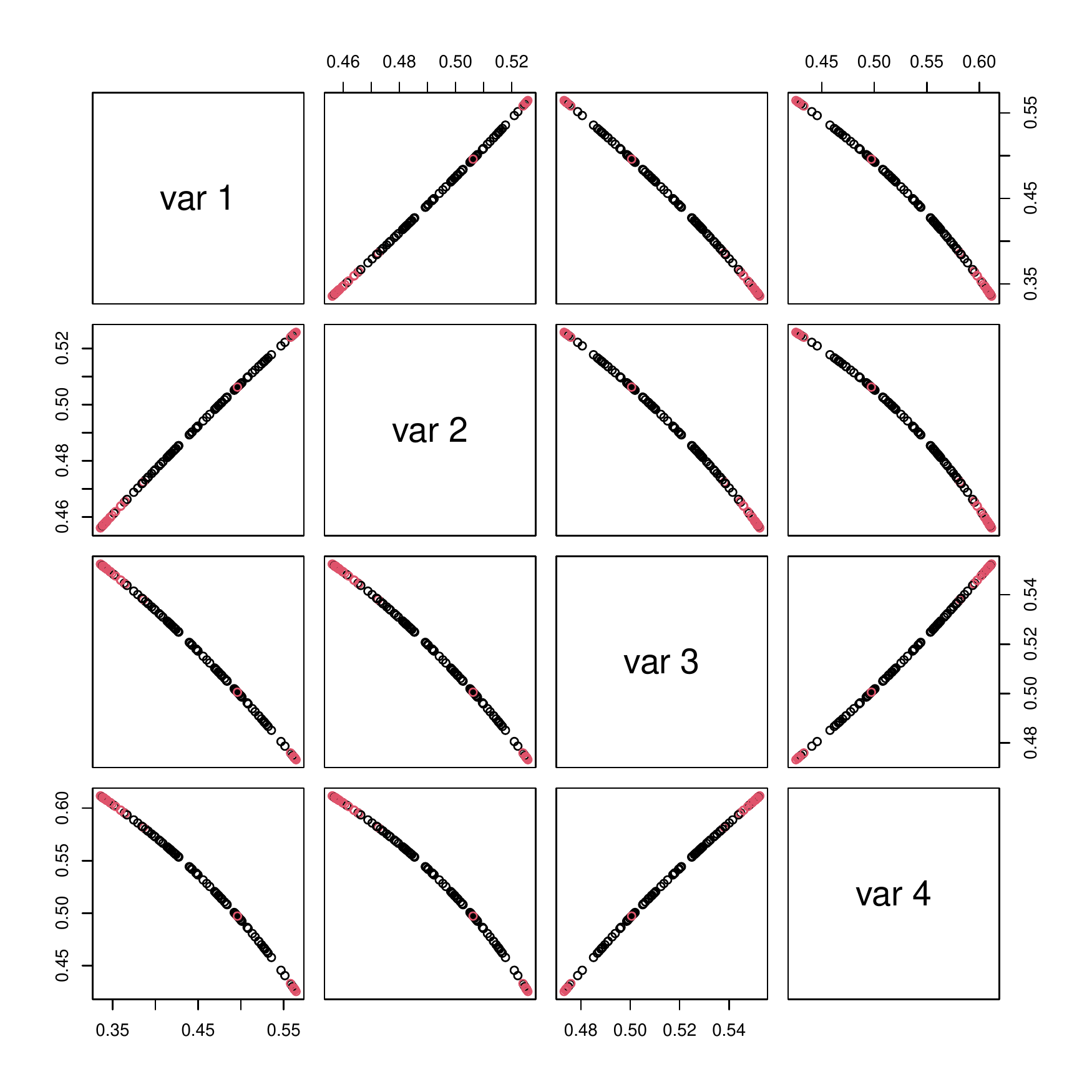}}

    \caption{{\small Pairwise scatterplots of the angular part of the extremes generated from a contaminated linear factor model and their corresponding kernel PCA preimages.  The red points denote extremes attributed to the signal $A \bZ_i$. The black points denote extremes attributed to the noise $\sigma\bm{\varepsilon}_i$.}}
    \label{fig:max-linear_data}
\end{figure}

As the screeplot in Figure \ref{fig:max-linear_screeplot} indicates, we use kernel PCA with $m=2$. Notice that the preimages of the kernel PCA shown on the right panel of Figure \ref{fig:max-linear_data} show the two-dimensional nature of the support of the 
 uncontaminated spectral measure much more clearly than the original extremes do. In particular,  we see clearly that the extremes generated from the linear factor model are mapped close to two points while most of the extremes due to the``noise'' are mapped in between these two points.

\newpage


 \subsection{Spiked angular Gaussian model}

We consider extremes arising from the model
\begin{equation*}
     \bX_i=  u_i\bN_i+\sigma\bm\varepsilon_i, \quad i=1,\dots,n,
 \end{equation*}
 where $\{u_i\}_{i=1}^n$ are  i.i.d.~univariate standard Fr{\'e}chet  random variables, $\{\bN_i\}_{i=1}^n$ are  i.i.d.~$d$-dimensional centered normal vectors with covariance matrix of the form   
\begin{equation}
\label{eq:spiked}
\Sigma=\sum_{k=1}^p\lambda_k\bv_k\bv_k^\top +\sigma_0^2 I_d
\end{equation} 
for $1\leq p \leq d$, 
where $\lambda_1\geq \lambda_2\geq \cdots \geq \lambda_p >0$ and the vectors  $\bv_1,\dots,\bv_p$ are orthonormal, 
and the terms $\{\sigma\bm\varepsilon_i\}_{i=1}^n$ are ``contamination'' terms of the same type as in the contaminated linear factor model example. The covariance matrix $\Sigma$ in \eqref{eq:spiked}  is a popular model that received a lot of attention in the machine learning and high dimensional (but non-extreme) statistics  in recent years.

We note that when $\sigma=0$, the spectral distribution on the $d$-dimensional sphere is given by 
\begin{equation}\label{eq:SAG_gamma}
\Gamma(\cdot)=\E\left(\|\bN\|\delta_{\frac{\bN}{\|\bN\|}}(\cdot)\right)/C\,,
\end{equation}
where $C=\E\|\bN\|$ (see equation (6.4) in \cite{avellamedinaetal2021}).  Assuming the model in \eqref{eq:spiked}, the spectral distribution $\Gamma$  has a spiked angular central Gaussian  distribution\footnote{Note that this is not exactly the same angular Gaussian distribution of \cite{tyler1987} because of the exponent $-(d+1)/2$ due to the presence of $\|\bN\|$ in \eqref{eq:SAG_gamma}. } on the $d$-dimensional sphere $\mathbb S^{d-1}$ with density function  given by
\begin{equation*}
\label{angGaussian}
g(\bomega;\Sigma)=C^{-1}\frac{2\pi^{d/2}}{\Gamma(d/2)}|\Sigma|^{-1/2}(\bomega^\top\Sigma^{-1} \bomega)^{-(d+1)/2}, \quad \bomega\in \mathbb S^{d-1}\,.
\end{equation*}
%
Intuitively, this model generates $r$ clusters of extremes corresponding to higher density regions of the angular Gaussian distribution given by the principal directions of $\Sigma$. 

In our experiment we take $d=4$
and $p=2$, and use $\Sigma=BB^\top$ where
 $$B^\top=
 \begin{pmatrix} 
 0.1 &0.2&0.3&0.4 \\
 0.9& 0.8& 0.7& 0.6
 \end{pmatrix},
$$
so that $\bv_1$ and $\bv_2$ are the left singular vectors of $B$. We have chosen   $\sigma=0.1$ and   $\sigma_0=1$.   
 %
 
\begin{figure}[h!]
    \centering
    \includegraphics[scale=0.35]{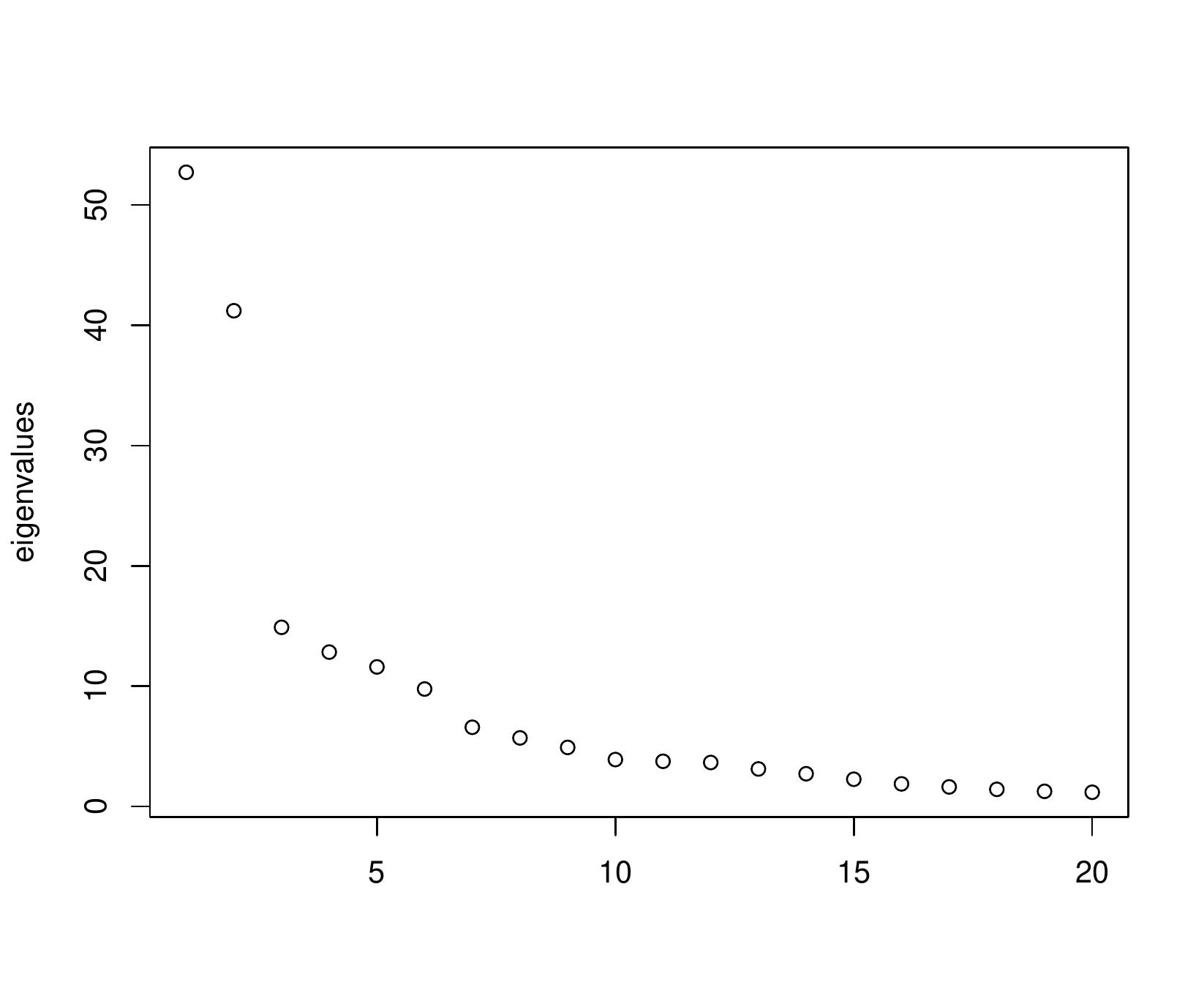}
    \caption{{\small Largest 20 eigenvalues of the kernel matrix used to run kernel PCA for the 4 dimensional contaminated spiked angular Gaussian model.}}
    \label{fig:spiked_screeplot}
\end{figure}

\begin{figure}[h!]
    \centering
    \subfigure[Contaminated spiked angular Gaussian data]{\includegraphics[scale=0.35]{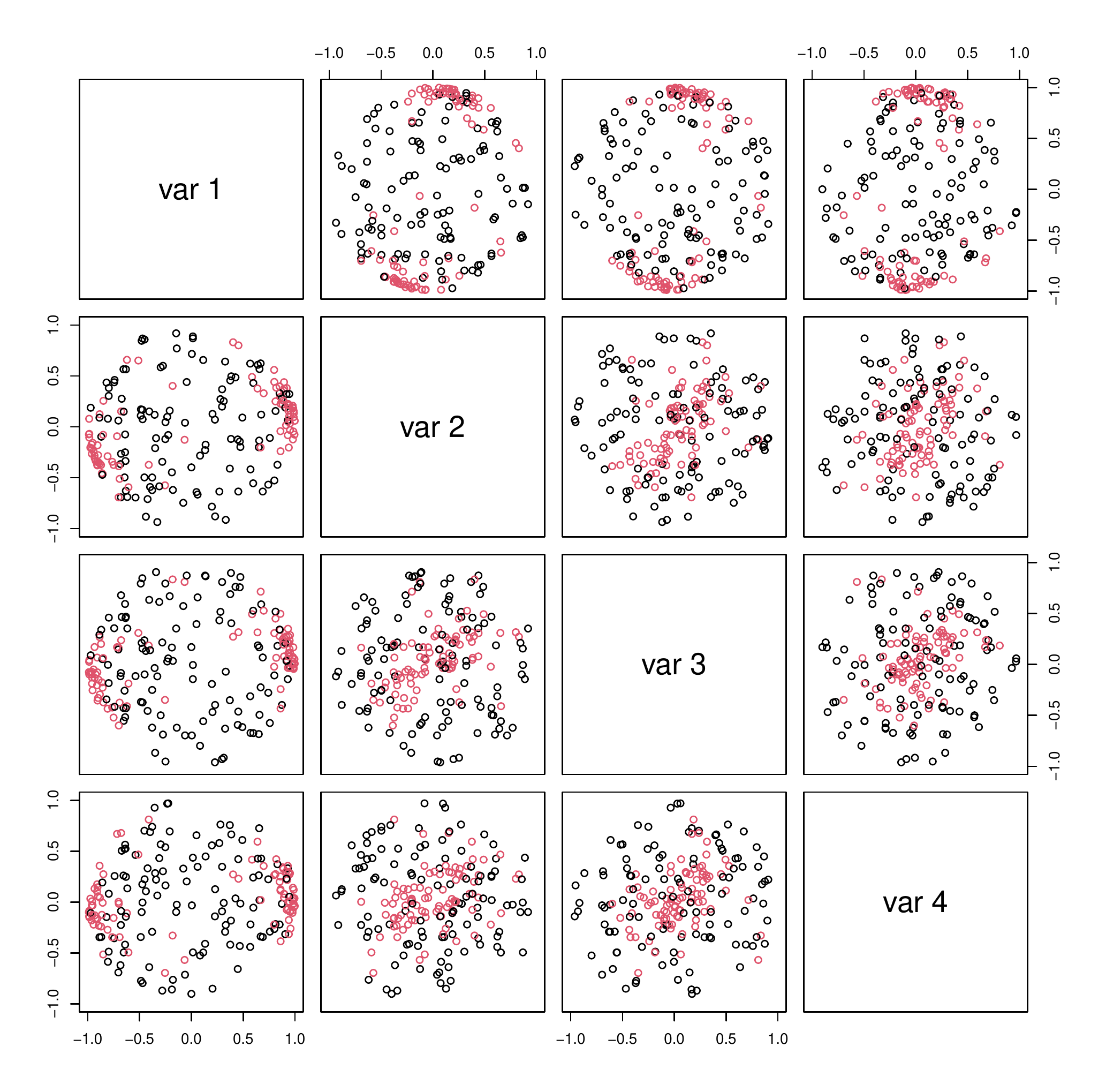}}
\hfill
\subfigure[Preimages, $\gamma=0.5$]{\includegraphics[scale=0.35]{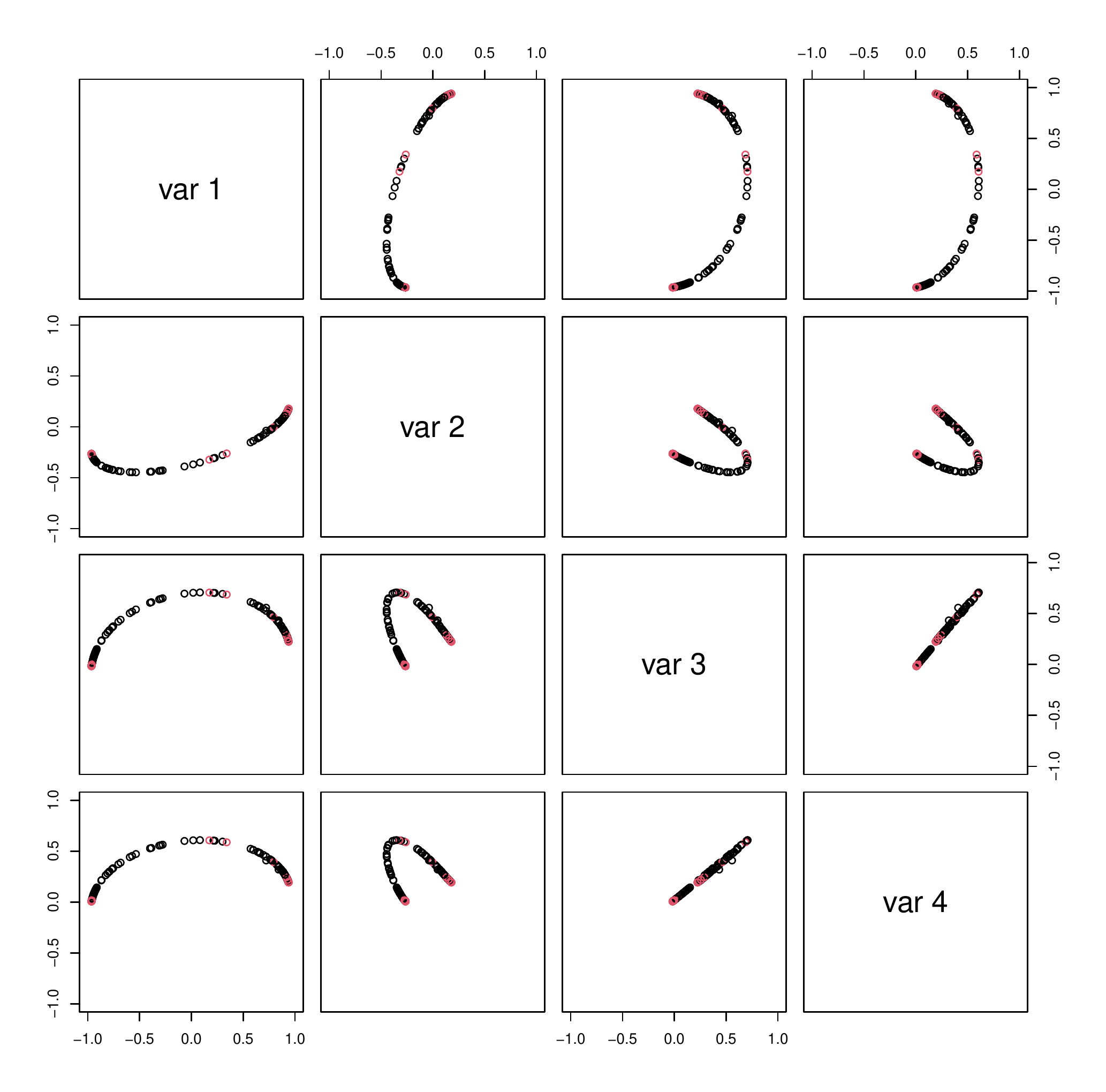}}
    \caption{{\small Pairwise scatterplots of the extremes generated from the spiked angular Gaussian model with $d=4$ and $r=2$.  The red points denote extremes attributed to the signal $u_i\bZ_i$. The black points denote extremes attributed to the noise $\sigma\varepsilon_i$.}}
    \label{fig:spiked_data}
\end{figure}

 According to the screeplot of Figure \ref{fig:spiked_screeplot} we use kernel PCA with $m=2$.  Once again, the dramatic dimension reduction of the support of the extremes after going through kernel PCA is clear in 
 Figure \ref{fig:spiked_data}.

%

%
%

\newpage

 \subsection{Approximate subspace model: regularly varying circle}

\begin{figure}[h!]
    \centering
    \subfigure[Noisy subspace data]{\includegraphics[scale=0.35]{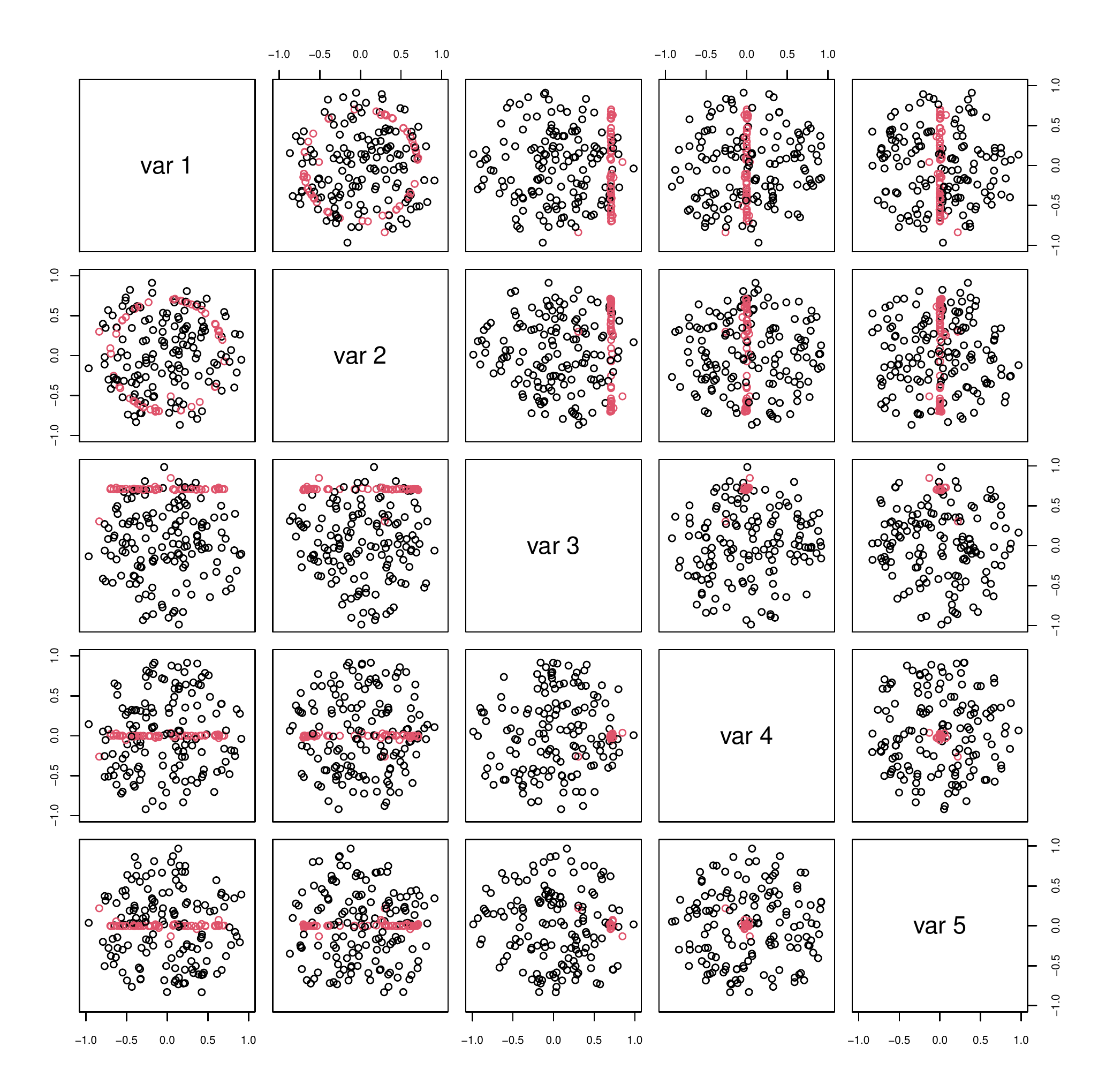}}
\hfill
\subfigure[Preimages]{\includegraphics[scale=0.35]{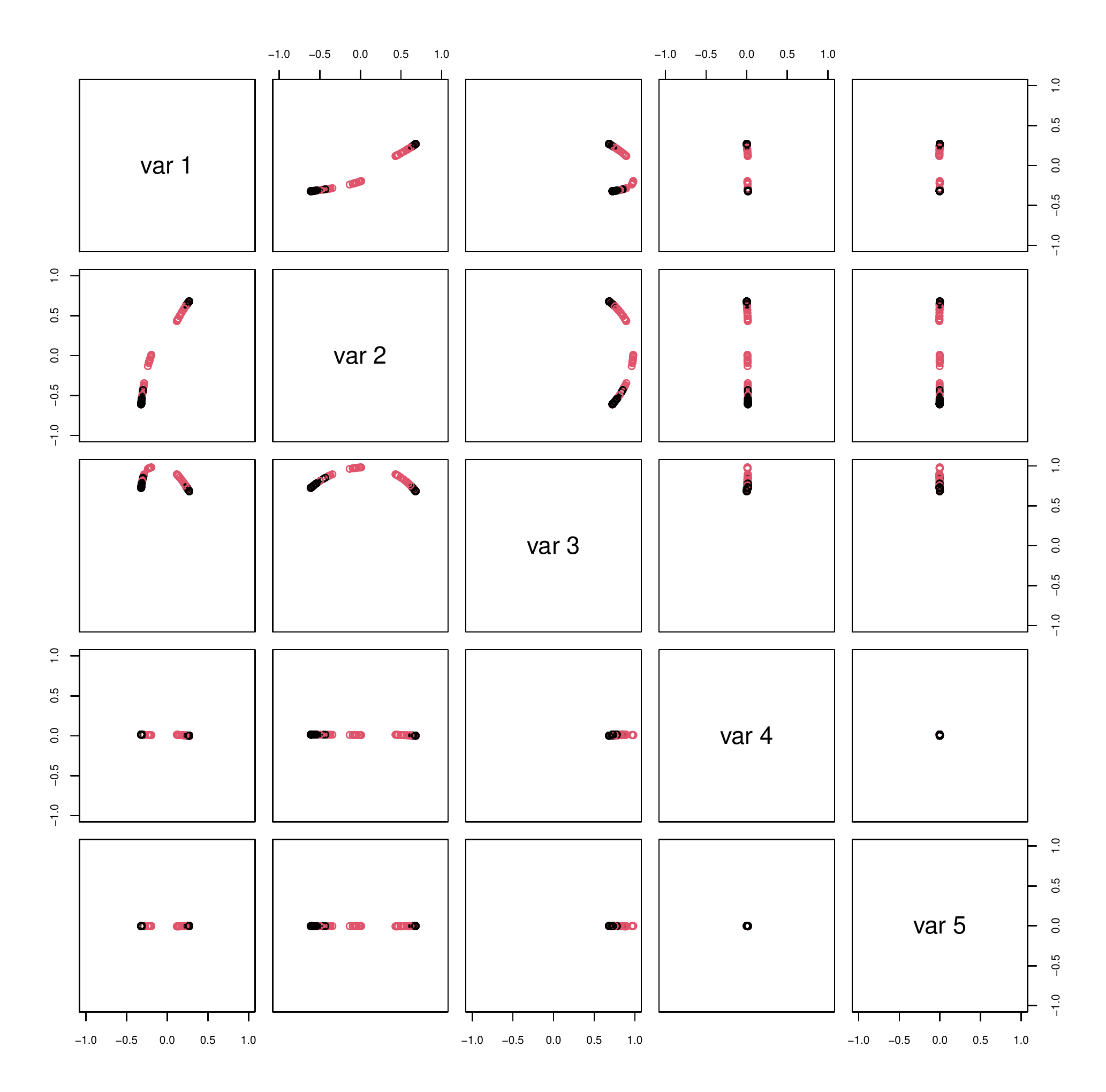}}
    \caption{{\small Pairwise scatterplots of the extremes generated from the noisy circle subspace model.  The red points denote extremes attributed to the subspace signal. The black points denote extremes attributed to the noise $\sigma\varepsilon_i$.}}
    \label{fig:circle_data}
\end{figure}

We also consider a model where the ambient dimension is 5 but the signal is driven by a 3-dimensional regularly varying vector with spectral measure supported on a circle. More specifically, we consider the model 
$$ \bX_i=\bZ_i+\sigma\bm{\varepsilon}_i, \quad i=1,\dots,n,$$
where $\bZ_i\in\mathbb{R}^5$ is such that its last 2 components are 0 and its first 3 entries are of the form 
$$Z_{i1}=Y_iG_{i1},\quad Z_{i1}=Y_iG_{i2},\quad Z_{i3}=Y_i\{(G_{i1}^2+G_{i2}^2)^{1/2}\},$$
where ${Y_i}$ is an i.i.d.~sequence of standard Fr{\'e}chet random variables and ${\bm{G}_i}$ is a sequence of bivariate standard i.i.d.~normal random variables. It follows that $\bZ_i$ is regularly varying, and its spectral measure is uniform on the circle $\{(z_1,z_2,z_3):z_1^2+z_2^2=1/2, \,z_3=1/\sqrt{2}\}$. The constant $\sigma>0$ regulates the signal to noise ratio and  $\bm\varepsilon_i$ is  noise vector obtained by multiplying a univariate independent  standard Fr{\'e}chet with an independent $p$-dimensional composed by the absolute value of i.i.d.~standard normals. In our example we chose $\sigma=2$ which leads to about about $60$ out of $200$ extremes generated from the signal of the circle. We see from Figure \ref{fig:circle_data} that very much like the last two examples, the kernel PCA preimage map the data to a lower dimensional subspace where distinguished the locations of the signal and the noise terms. In particular, we observe that the preimages correctly identify the structure of the subspace, mapping to variables 4 and 5 collapse to one point and identifying the straight lines of the signal as seen in the original colored data points. The subspace corresponding to the circle is distorted but the method recognizes seem to recognize that there is lower dimensional structure in the first 3 coordinates.


\subsection{Extremes from time series: ARCH(1) process}

\begin{figure}[h!]
    \centering
    \includegraphics[scale=0.42]{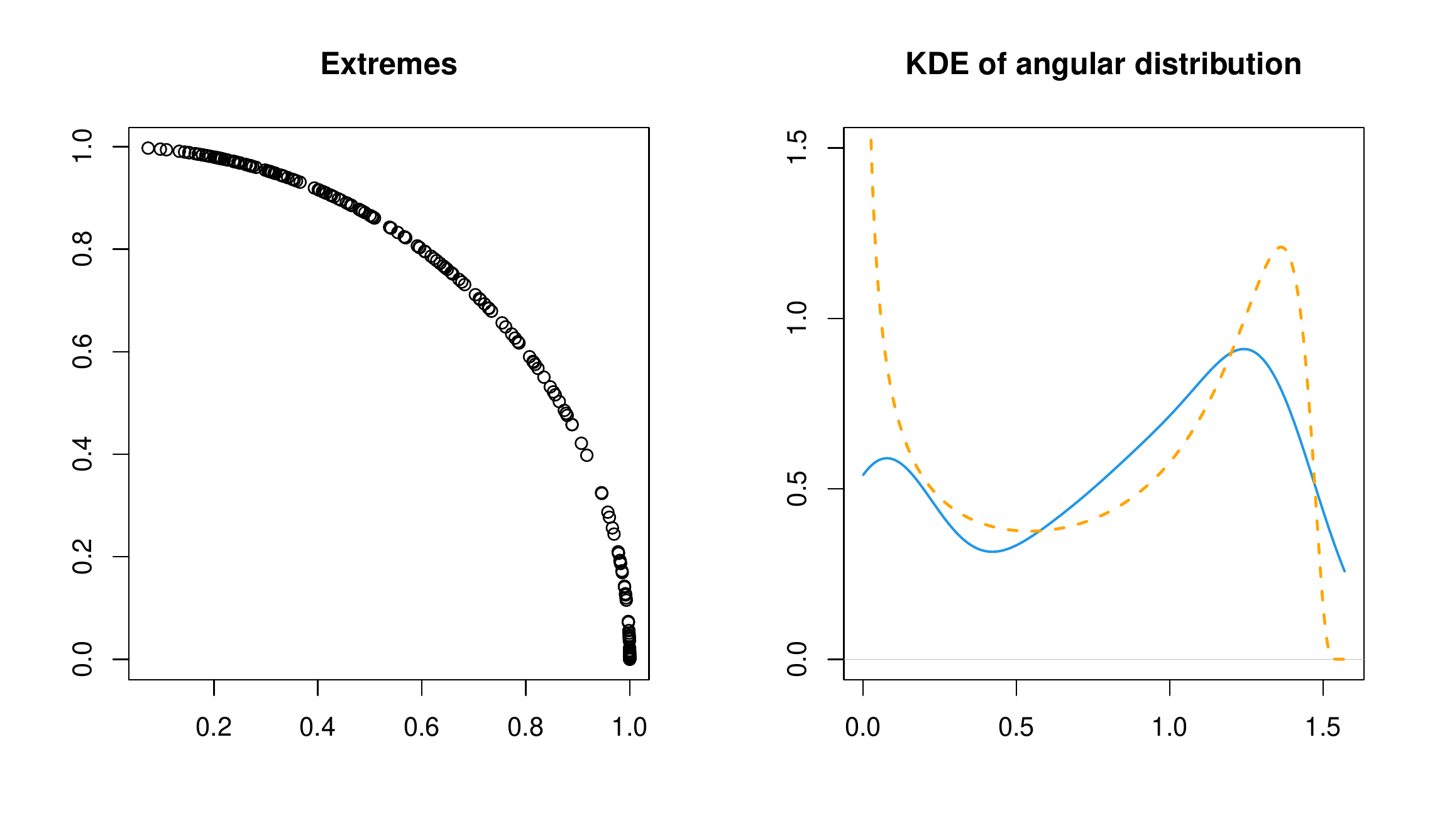}
    \caption{{\small The plot on the left shows the empirical sample of extremes $\bX_t/\|\bX_t\|$. The right hand side shows two density functions. The solid blue line is the kernel density estimator of the angles of the extremes fitted with default values of the \texttt{R} function \texttt{density}$(\cdot)$ i.e., using the Gaussian kernel with Silverman's rule of thumb. The dashed orange line corresponds to the theoretical spectral density.}}
    \label{fig:ARCH1_data}
\end{figure}

\begin{figure}[h!]
    \centering
    \includegraphics[scale=0.42]{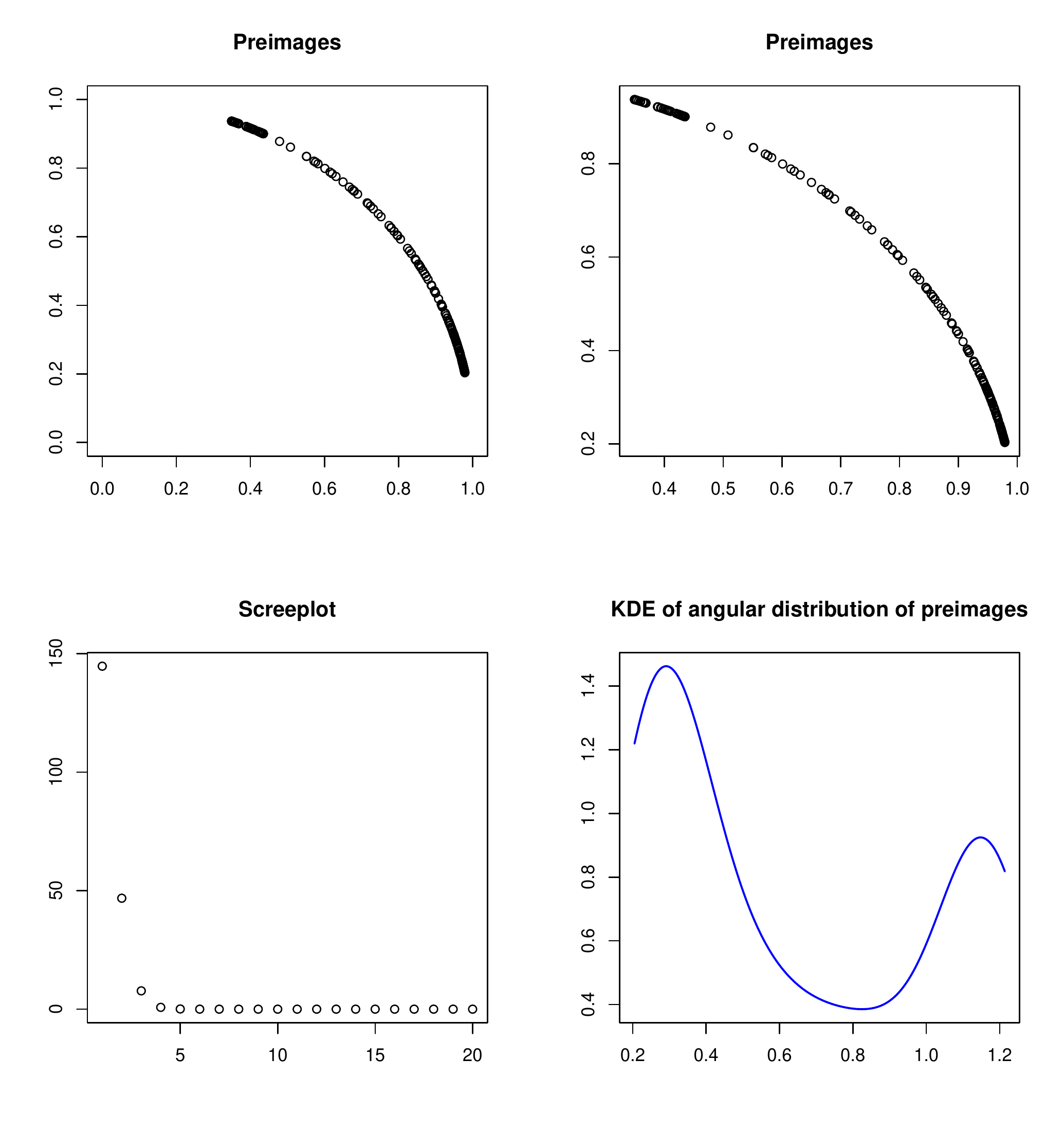}
    \caption{{\small The top plots show the kernel PCA preimages in the original scale of the data and on the scale of the preimages. The bottom plot shows the screeplot of the kernel matrix and the kernel density estimator of the angle of the preimages fitted with default values of the \texttt{R} function \texttt{density}$(\cdot)$.}}
    \label{fig:ARCH1_preimages}
\end{figure}

 When the extremes arise from a time series model, the independence assumption is, generally, violated.  Here we consider the square of a standard {\it integrated} ARCH(1) process that  follows the recursions,  $Y_t=(1+Y_{t-1})Z_t^2$,
 where  $\{Z_t\}$ is an i.i.d. sequence of standard Gaussian random variables. There is a unique stationary solution $\{Y_t\}$ to these recursions such that $Y_t$ has asymptotically Pareto tails with $\alpha=1$ (see \cite{basrak2002regular} and \cite{basrak2002characterization} for more details).  Here we consider the two-dimensional vector $\bX_t=(Y_{t-1},Y_{t})^\top$, whose tails are asymptotically equivalent to those of  the vector $(1, Z_t^2)^\top Y_{t-1}$. By an application of Breiman's lemma \cite[Proposition A.1]{basrak2002regular}, the spectral distribution of this vector on $[0,\pi/2]$ is given by
 $$
 \Gamma(\cdot)=\E\left(|1+Z_t^4|^{1/2}\delta_{\arctan(Z_t^2)}(\cdot)\right)/\E|1+Z_t^4|^{1/2}\,.
 $$
 The spectral measure can be shown to be bimodal as is also suggested by a kernel density estimator obtained from the angles of the empirical sample of extremes displayed in Figure \ref{fig:ARCH1_data}. 
 We look for clusters in the extremes of $\|\bX_t\|_2$ and note that the spectral measure of this example is continuous and hence it is less clear what the correct number of clusters should be. The preimages shown in  Figure \ref{fig:ARCH1_preimages} where obtained with kernel PCA with $m=3$, as suggested by the screeplot. The bottom right plot displays reveals that the distribution of the angles of the preimages remains bimodal with slightly more pronounced modes with more mass in between them. This is also reflected in the top right plot where one can perceive a third cluster of preimages.

 \newpage

\section*{Appendix}
				
				\subsection*{ Proof of Theorem \ref{thm:FrobScen1}}
				
				We start by considering the ``diagonal terms'' in
				\eqref{e:Frob.B} and note that for $k=1,\ldots, p$, 
				\begin{align*} 
					F_{k,k}(n) &=\frac{1}{N_n^2} \sum_{i\in \calI^{(k)}_n}
					\sum_{j\in \calI^{(k)}_n}
					\Bigl[R\bigl(\BX_i/ \|\BX_i\|-\BX_j/ 
					\|\BX_j\|\bigr)-R(0)\Bigr]^2 \\
					\notag &=  \frac{1}{N_n^2} \sum_{i=1}^{N_n^{(k)}}
					\sum_{j=1}^{N_n^{(k)}}
					\Bigl[R\bigl(\BY_i^{(k)}-\BY_j^{(k)}\bigr)-R(0)\Bigr]^2
				=\frac{(N_n^{(k)})^2}{N_n^2} G_{k,k}(n)\,.
				\end{align*}
				Recall that the law of each $\BY_i^{(k)},\,i=1,\ldots,N_n^{(k)}$ 
				is the conditional law of
				$\BX/\|\BX\|$ given $\| \BX\|>u_n, \,
				Z_{k}>u_n/w^{1/\alpha}$, different $\BY_i^{(k)}$ are independent, and also
				independent of $N_n$ and $N_n^{(k)}$.  
				For a large $M>0$ write
				\begin{align*} \notag
					G_{k,k}(n) =&  
					\frac{1}{(N_n^{(k)})^2} \sum_{i=1}^{N_n^{(k)}}
					\sum_{j=1}^{N_n^{(k)}}
			\left[R\left(\BY_i^{(k)}-\BY_j^{(k)}\right)-R(0)\right]^2 \one\left(\left\|\BY_i^{(k)}-\BY_j^{(k)}\right\|>Mu_n^{-1}\right)
					\\
				+& 
					\frac{1}{(N_n^{(k)})^2} \sum_{i=1}^{N_n^{(k)}}
					\sum_{j=1}^{N_n^{(k)}}
					\left[R\left(\BY_i^{(k)}-\BY_j^{(k)}\right)-R(0)\right]^2 \one\left(\left\|\BY_i^{(k)}-\BY_j^{(k)}\right\|\leq Mu_n^{-1}\right)\\
					\notag =:& G_{k,k}^{>M}(n)+ G_{k,k}^{\leq M}(n). 
				\end{align*}
				Clearly,
				\begin{align*} 
					& \E[ u_n^{2\theta}  G_{k,k}^{>M}(n)] 
					\\ \notag & \hskip 0.2 in \leq 
					u_n^{2\theta} \E\left\{ \left[R\left(\BY_1^{(k)}-\BY_2^{(k)}\right)-R(0)\right]^2 \one\left(\left\|\BY_1^{(k)}-\BY_2^{(k)}\right\|>Mu_n^{-1}\right)\right\}.
				\end{align*}
				If we can show that
				\begin{equation} \label{e:unif.int}
					\sup_{n} u_n^{(2+\vep)\theta}\E \Bigl[R\bigl(\BY_1^{(k)}-\BY_2^{(k)}\bigr)-R(0)\Bigr]^{2+\vep}<\infty
				\end{equation}
				for some $\vep>0$ (in which case,  $u_n^{2\theta}\left(R(\BY_1^{(k)}-\BY_2^{(k)})-R(0)\right)^2$ is uniformly integrable), it will follow from Theorem \ref{pr:Delta.B} that  
				\begin{equation} \label{e:first.m.1}
					\limsup_{n\to\infty} \E\left[ u_n^{2\theta}  G_{k,k}^{>M}(n)\right]\leq d_\theta^2
					\E\Bigl\{ \bigl\|\bS_1^{(k)}-\bS_2^{(k)}\bigr\|^{2\theta} \one\bigl(
					\bigl\|\bS_1^{(k)}-\bS_2^{(k)}\bigr\|> M/d_\theta\bigr)\Bigr\}. 
				\end{equation}
				By \eqref{e:cov.zero} the bound \eqref{e:unif.int} will follow once we
				check that
				\begin{equation} \label{e:unif.int.1}
					\sup_{n} u_n^{(2+\vep)\theta}\E \left[\Bigl\|\BY_1^{(k)}-\bs_k\Bigr\|^{(2+\vep)\theta}\right]<\infty. 
				\end{equation}
				Using the notation in (4.13) and (4.14) in \cite{avellamedinaetal2021} it suffices to check 
					that for any $l=1,\ldots, d$,
					$$
					\sup_n \E\left[\left| w_j^2\left(\sum_{m=1}^p a_{lm}Z_m\right)^2-a_{lj}^2 \|\BX\|^2
					\right| ^{(2+\vep)\theta}\right]<\infty. 
					$$
					The expectation, however, is independent of $n$ and is finite  for $\epsilon>0$ sufficiently small since $\alpha>2\theta$ by assumption.  This establishes 
				\eqref{e:first.m.1}, whence
				\begin{equation} \label{e:first.m.2}
					\lim_{M\to\infty}  \limsup_{n\to\infty} \E \left[u_n^{2\theta}  G_{k,k}^{>M}(n)\right]=0.
				\end{equation}
				The same argument shows that for every fixed $M>0$, 
				\begin{equation} \label{e:first.m.3}
					\lim_{n\to\infty} \E u_n^{2\theta}  G_{k,k}^{\leq M}(n)= d_\theta^2
					E\Bigl\{ \bigl\|\bS_1^{(k)}-\bS_2^{(k)}\bigr\|^{2\theta} \one\bigl(
					\bigl\|\bS_1^{(k)}-\bS_2^{(k)}\bigr\|\leq M/d_\theta\bigr)\Bigr\}. 
				\end{equation}
				Furthermore,
				\begin{align} 
				\notag	\E \bigl(u_n^{2\theta}  G_{k,k}^{\leq M}(n)\bigr)^2
					=&  \E\left( \frac{N_n^{(k)} -1}{(N_n^{(k)})^3}\one\left(N_n^{(k)}\ge1\right)\right) E_n^{(1)} +
					4\E\left( \frac{(N_n^{(k)} -1)^2}{(N_n^{(k)})^3}\one\left(N_n^{(k)}\ge1\right)\right) E_n^{(2)} \\
					 &  +\E\left( \frac{(N_n^{(k)} -1) (N_n^{(k)} -2) (N_n^{(k)} -3)
					}{(N_n^{(k)})^3}\one\left(N_n^{(k)}\ge1\right)\right) E_n^{(3)}, \label{e:first.m.4}
				\end{align}
				where 
				\begin{align*}
					&E_n^{(1)}= \E\left\{ u_n^{4\theta} \left[R\left(\BY_1^{(k)}-\BY_2^{(k)}\right)-R(0)\right]^4 \one\left(\left\|\BY_1^{(k)}-\BY_2^{(k)}\right\|\leq Mu_n^{-1}\right)\right\} \\
					&E_n^{(2)}= \E\left\{ u_n^{4\theta} \left[R\left(\BY_1^{(k)}-\BY_2^{(k)}\right)-R(0)\right]^2
					\left[R\left(\BY_1^{(k)}-\BY_3^{(k)}\right)-R(0)\right]^2 \right. \\
					&\left. \hskip 1in 
					\cdot\one\left(\left\|\BY_1^{(k)}-\BY_2^{(k)}\right\|\leq Mu_n^{-1}, \ 
					\left\|\BY_1^{(k)}-\BY_3^{(k)}\right\|\leq Mu_n^{-1}\right)\right\} \\
					&E_n^{(3)}= \left(\E\left\{ u_n^{4\theta} \left[R\left(\BY_1^{(k)}-\BY_2^{(k)}\right)-R(0)\right]^2 \one\left(\left\|\BY_1^{(k)}-\BY_2^{(k)}\right\|\leq Mu_n^{-1}\right)\right\}\right)^2. 
				\end{align*}
				It follows from \eqref{e:cov.zero} that, for a fixed $M$,
				$E_n^{(1)}$ and $E_n^{(2)}$ are bounded by an $M$-dependent constant,
				so the first two terms in the right hand side of \eqref{e:first.m.4}
				vanish in the limit. Furthermore, it follows by \eqref{e:unif.int}
				that
				$$
				E_n^{(3)}\to  \biggl(d_\theta^2
				\E\Bigl\{ \bigl\|\bS_1^{(k)}-\bS_2^{(k)}\bigr\|^{2\theta} \one\bigl(
				\bigl\|\bS_1^{(k)}-\bS_2^{(k)}\bigr\|\leq M/d_\theta\bigr)\Bigr\}\biggr)^2. 
				$$
				Therefore,
				$$
				\lim_{n\to\infty} \E \bigl(u_n^{2\theta}  G_{k,k}^{\leq M}(n)\bigr)^2=
				\bigl(\lim_{n\to\infty} \E u_n^{2\theta}  G_{k,k}^{\leq M}(n)\bigr)^2
				$$
				and consequently
				\begin{equation} \label{e:first.m.5}
					\lim_{n\to\infty}\var \bigl(u_n^{2\theta}  G_{k,k}^{\leq M}(n)\bigr)= 0
				\end{equation}
				It follows from \eqref{e:first.m.3} and \eqref{e:first.m.5} that
				\begin{equation} \label{e:first.m.6}
					u_n^{2\theta}  G_{k,k}^{\leq M}(n) \cip
					d_\theta^2
					\E\Bigl\{ \bigl\|\bS_1^{(k)}-\bS_2^{(k)}\bigr\|^{2\theta} \one\bigl(
					\bigl\|\bS_1^{(k)}-\bS_2^{(k)}\bigr\|\leq M/d_\theta\bigr)\Bigr\}, \mbox{ as $n\to\infty$.}
				\end{equation}
				Now \eqref{e:kk.scen1} follows easily from \eqref{e:first.m.2}   and
				\eqref{e:first.m.6}. It follows from \eqref{e:kk.scen1} and \eqref{e:N.n}
				that
				\begin{equation} \label{e:kk.scen1a}
					u_n^{2\theta}  F_{k,k}(n) \cip  \left(\frac{d_\theta w_k^\alpha}{w}\right)^2
					\E\bigl\|\bS_1^{(k)}-\bS_2^{(k)}\bigr\|^{2\theta}, \mbox{ as $n\to\infty$.}
				\end{equation}

				We now consider the ``off-diagonal terms'' in
				\eqref{e:Frob.B} as described in \eqref{e:F.k1k2}. 
				We claim that, under \eqref{e:alpha.theta.1}, 
				\begin{equation} \label{e:k1k2.scen1}
					u_n^2 G_{k_1,k_2}(n) \cip \E\bigl[ \nabla
					R(\bs_{k_1}-\bs_{k_2})\bigl(\bS^{(k_1)}_1-\bS^{(k_2)}_1\bigr)\bigr]^2, \mbox{ as $n\to\infty.$}
				\end{equation}
				The argument is similar to the argument we used to
				prove \eqref{e:kk.scen1}. Once again we write for a large $M>0$
				\begin{align} \notag
					G_{k_1,k_2}(n) =&  
					\frac{1}{N_n^{(k_1)}N_n^{(k_2)}} \sum_{i=1}^{N_n^{(k_1)}}
					\sum_{j=1}^{N_n^{(k_2)}}
					\left[R\left(\BY_i^{(k_1)}-\BY_j^{(k_2)}\right)-R(\bs_{k_1}-\bs_{k_2})\right]^2 \\
					\notag & \hskip 2in 
					\cdot\one\left(\left\|\BY_i^{(k_1)}-\BY_j^{(k_2)}-(\bs_{k_1}-\bs_{k_2})\right\|>Mu_n^{-1}\right)
					\\
					\notag+& 
					\frac{1}{N_n^{(k_1)}N_n^{(k_2)}} \sum_{i=1}^{N_n^{(k_1)}}
					\sum_{j=1}^{N_n^{(k_2)}}
					\left[R\left(\BY_i^{(k_1)}-\BY_j^{(k_2)}\right)-R(\bs_{k_1}-\bs_{k_2})\right]^2 \\
					\notag & \hskip 2in 
					\cdot\one\left(\left\|\BY_i^{(k_1)}-\BY_j^{(k_2)}\right\|\leq Mu_n^{-1}\right) \\
					\label{e:split.M12} =:& G_{k_1,k_2}^{>M}(n)+ G_{k_1,k_2}^{\leq M}(n), 
				\end{align}
				and
				\begin{align*} 
					\notag & \E \left[u_n^{2}  G_{k_1,k_2}^{>M}(n)\right] \\ 
					& =
					u_n^{2} \E\left\{  
					\left[R\left(\BY_1^{(k_1)}-\BY_2^{(k_2)}\right)-R(\bs_{k_1}-\bs_{k_2})\right]^2 
					\cdot\one\left(\left\|\BY_1^{(k_1)}-\BY_2^{(k_2)}\right\|>Mu_n^{-1}\right)
					\right\}.
				\end{align*}
				By Theorem \ref{pr:Delta.B}, if we can show that for
				some $\vep>0$, 
				\begin{equation} \label{e:unif.int12}
					\sup_{n} u_n^{2+\vep}
					\E
					\left|R\left(\BY_1^{(k_1)}-\BY_2^{(k_2)}\right)-R(\bs_{k_1}-\bs_{k_2})\right|^{2+\vep}<\infty,
				\end{equation}
				then it would follow that
				\begin{equation} \label{e:first.m.12}
					\lim_{n\to\infty} \E\left[ u_n^{2}  G_{k_1,k_2}^{>M}(n)\right]= 
					\E\bigl[ \nabla
					R(\bs_{k_1}-\bs_{k_2})\bigl(\bS^{(k_1)}_1-\bS^{(k_2)}_2\bigr)\one\bigl( \|
					\bS^{(k_1)}_1-\bS_2^{(k_2)}\| >M/d_\theta\bigr) 
					\bigr]^2. 
				\end{equation}
				However, because of the assumed continuous differentiability of $R$
				outside of the origin, \eqref{e:unif.int12} follows immediately from
				\eqref{e:unif.int.1}. Therefore, \eqref{e:first.m.12} holds. The rest
				of the argument for \eqref{e:k1k2.scen1} 
				is the same as for the ``diagonal terms'', and it follows from
				\eqref{e:k1k2.scen1} that 
				\begin{equation} \label{e:k1k2.scen1f}
					u_n^2 F_{k_1,k_2}(n) \cip \frac{w_{k_1}^\alpha 
						w_{k_2}^\alpha}{w^2} \E\bigl[ \nabla
					R(\bs_{k_1}-\bs_{k_2})\bigl(\bS^{(k_1)}_1-\bS^{(k_2)}_2\bigr)\bigr]^2, \mbox{ as $n\to\infty.$} 
				\end{equation}
				This completes the proof. $\Box$

				\subsection*{Proof of Theorem \ref{pr:point_process}}
				
			 In order to establish the convergence in  \eqref{e:pp.conv}, it suffices to show that the corresponding Laplace functionals converge.  That is, it is enough to prove that 
					\begin{equation}
		\mathbb E\left[\exp\{-M_n(f)\}\right]\to 	\mathbb E\left[\exp\{-M_\alpha(f)\}\right]\,,
				\end{equation}
				for any nonnegative, bounded continuous function $f$ with support outside a neighborhood of the origin. Here $M(f)$ is shorthand for $\int_{\RR^d\setminus \{0\}} f(\bx)M(d\bx)$; 
				see Theorem 5.1 in \cite{resnick2007} (this book can be consulted also for more  details on Poisson processes and weak convergence to  point processes). 
				
				Adapting Theorem 5.3 in \cite{resnick2007} to random sums of point measures and using the fact that 
				$$
				N_n^{(k)} \sim c_\alpha w_k^{\alpha} u_n^{-\alpha} n \ \ \mbox{in probability as $n\to\infty.$},
				$$
				it suffices to show that the intensity measures converges vaguely, i.e.,
					\begin{equation} \label{e:mean4}
						c_\alpha w_k^{\alpha} u_n^{-\alpha}n \bbP\bigl(u_n^2 n^{-1/\alpha} 
						\BY_0\bigr)\in \cdot \Bigr) \civ m_\alpha^{(k)}(\cdot)\,,
					\end{equation}
					where $\civ$ denotes vague convergence of measures on $\bbr^p\setminus\{0\}$ and $\BY_0=\BY_1^{(k)}-\bs_k$.  To ease notation we write $\bbP_{k,n}(\cdot)=\bbP\left(\,\cdot\, \,|\, \|\BX\|>u_n, Z_k>u_n/ w^{1/\alpha} \right)$ and $\E_{k,n}$ to be the corresponding conditional  expectation.
				We start by showing  that 
					\begin{eqnarray}\label{e:firststep}
						\nu_n(A\times B)&:=&c_\alpha w_k^{\blue {\alpha}} u_n^{-\alpha}n\bbP_{k,n}\left( (u_nn^{-1/\alpha}\bZ_{-k}, u_n^{-1}Z_k)\in A\times B \right) \nonumber\\
						&\civ& 
						c_\alpha w_k^{\blue {\alpha}}\nu(A) \mu(B)\,, 
					\end{eqnarray}
					for all {\it bounded Borel sets}   $A\times B\subset ([0,\infty)^{p-1}\setminus \{0\})\times [w_k^{-1},\infty)$ that are also continuity sets of the limit measure,  where $\nu$ is the measure on $[0,\infty]^{p-1}\setminus\{0\}$ given by
					$$
					\nu(A)=\sum_{j\ne k} \int_{0}^\infty \one_{A^{(j)}}(x)\alpha x^{-\alpha-1}\,dx\,,
					$$
					$\bZ_{-k}$ is the vector $(Z_1\ldots,Z_p)^\top$ with the $k$th component omitted, $A^{(j)}$ is the intersection of the set $A$ with the $j$th coordinate axis, and $\mu(x,\infty)=w_k^{-\alpha} x^{-\alpha}$, $x\ge w_k^{-1}$.  
					To prove \eqref{e:firststep}, consider $j=1$, $A=(x,\infty)\times \bbr^{p-2}$, $B=(y,\infty)$ with $y\ge w_k^{-1}$ and $k\not=1$. 
					Then 
					\begin{eqnarray*}
						\nu_n(A\times B)&=&c_\alpha w_k^ {\alpha}u_n^{-\alpha}n\bbP_{k,n}\left( u_nn^{-1/\alpha}Z_1>x, u_n^{-1}Z_k>y\right)\\
						&\sim&c_\alpha w_k^{\alpha} u_n^{-\alpha}nc_\alpha x^{-\alpha}u_n^{\alpha}n^{-1}w_k^{-\alpha} y^{-\alpha}\\
						&\to& c_\alpha^2 w_k^{\alpha} x^{-\alpha}w_k^{-\alpha} y^{-\alpha}\\
						&=&c_\alpha^2 w_k^{\alpha}\nu(A)\mu(B)\,.
					\end{eqnarray*}
					The argument for general choices of sets $A$ and $B$ is similar.

We now show  how to derive \eqref{e:mean4} from \eqref{e:firststep} using the continuous mapping theorem.  On the event $\|\BX\|>u_n, Z_k>u_n/w^{1/\alpha}$, it follows that $\|\BX\| \sim w_k |Z_k|$, so the expression in (4.13)  of \cite{avellamedinaetal2021} corresponding to $(V_1,\ldots,V_p)^\top=u_n\BY_0$ can be written as
\begin{equation}\label{e:Vl}
	V_l =u_n\frac{2a_{lk}Z_kT_{lk} +w_k X^2_{l,-k} - a_{lk}^2
	\sum_{i=1}^d  X^2_{i,-k}}{2w_k^2 a_{lk}Z_k^2}(1+o(1))\,,
\end{equation}
where
\begin{eqnarray*}\label{e:T}
		T_{lk} &=& w_k^2 X_{l,-k} -a_{lk} \sum_{i=1}^d a_{ik}X_{i,-k}\\
		&=&w_k^2\sum_{j\ne k}b_l^{(j,k)}Z_j\,,
\end{eqnarray*}
and
\begin{equation*} \label{e:b.hat}
\hat b_l^{(j,k)} = a_{lj} -\frac{a_{lk}}{w_k^2}\sum_{i=1}^d
a_{ik}a_{ij}\,.
\end{equation*}
Note that for any $i=1,\ldots, d$, and $c>0$  a constant that may change from line to line, we have  by \eqref{e:patt.i}  below, with $z=cn^{1/\alpha}u_n^{-1}$,
\begin{eqnarray*}
u_n^{-\alpha}n\bbP_{k,n}\left(u_n\frac{X^2_{i,-k}}{Z_k^2}>cn^{1/\alpha}u_n^{-1}\right) 
&\leq&C u_n^{-\alpha}n u_n^{-\alpha/2}n^{-1/2}u_n^{\alpha/2} \\
& \sim& C u_n^{-\alpha} n^{1/2} 
\to 0
\end{eqnarray*}
as $n\to\infty$ by \eqref{e:level.unusual}. Therefore, writing 
$\tilde \BV=(\tilde V_1,\ldots, \tilde V_l)^T$ with 
\begin{equation*} \label{e:V.tilde}
\tilde V_l= \frac{u_n  T_{lk}}{w_k^2Z_k}=u_n\frac{\sum_{j\ne k}\hat b_l^{(j,k)}Z_j}{Z_k}, \ \ l=1,\ldots, d, 
\end{equation*}
we have 
\begin{equation} \label{e:mean}
	c_\alpha w_k^{\alpha} u_n^{-\alpha}n \bbP\bigl(u_n^2 n^{-1/\alpha} 
	\BY_0\in \cdot \bigr)=c_\alpha w_k^{\alpha}u_n^{-\alpha}n\bbP_{k,n}(u_n n^{-1/\alpha} 
	\tilde{\BV}\in \cdot)+o(1)\,.
\end{equation}
So to finish the proof, it suffices to show
\begin{equation} \label{e:mean3}
	c_\alpha w_k^{\alpha}u_n^{-\alpha}n \bbP_{k,n}(u_n n^{-1/\alpha} 
	\tilde\BV\in \cdot)\civ m_\alpha^{(k)}(\cdot)\,.
\end{equation}
Consider the continuous mapping from $E:=([0,\infty)^{p-1}\setminus\{0\})\times [w_k^{-1},\infty)$  to $\bbr^{d}\setminus\{0\}$ given by
\begin{eqnarray*}
	T(z_{-k},y)&=&\left(\sum_{j\ne k}\hat b_l^{(j,k)}z_j/y\,,l=1,\ldots,d\right)\\
	&=&\sum_{j\ne k}\hat \boldb^{(j,k)}z_j/y\,,
\end{eqnarray*}
where $\hat \boldb^{(j,k)}= \bigl(  \hat b_1^{(j,k)},\ldots, \hat
b_d^{(j,k)}\bigr)^\top$.
	Since this mapping has the property that for a compact set $K$ in $E$, $T^{-1}(K)$ is also compact	in $E$, it follows from Proposition 5.2 in \cite{resnick2007} that for any Borel set $A$ in $\bbr^d$ bounded away from $0$  with  $\nu\times\mu( \partial (T^{-1}(A))=0$, the left hand side of \eqref{e:mean3} is
\begin{eqnarray*}
	\nu_n\circ T^{-1}(A)&=&c_\alpha w_k^{\alpha} u_n^{-\alpha}n	\bbP_{k,n}\left(T(u_nn^{-1/\alpha}\bZ_{-k}, u_n^{-1}Z_k)\in A\right)\\
	&=&c_\alpha w_k^{\alpha} u_n^{-\alpha}n\bbP_{k,n}\left(\frac{u_nn^{-1/\alpha}\sum_{j\ne k} \hat \boldb^{(j,k)}Z_{j}}{u_n^{-1}Z_k} \in A\right)\\
	&\to&c_\alpha^2 w_k^{\alpha}(\nu\times \mu) \circ T^{-1}(A)\,.
\end{eqnarray*}
Finally,   by Fubini's theorem, 
\begin{eqnarray*}
c_\alpha^2 w_k^{\alpha}(\nu\times \mu) \circ T^{-1}(\cdot)&=&
c_\alpha^2 w_k^{\alpha}\sum_{j\ne k}\int_{w_k^{-1}}^\infty\left(\int_{0}^\infty \alpha z^{-\alpha-1}\delta_{\hat \boldb^{(j,k)}z/y}(\cdot)\,dz\right)\,\alpha w_k^{-\alpha} y^{-\alpha-1}dy\,,\\
&=&
c_\alpha^2 w_k^{\alpha}\sum_{j\ne k}\int_{0}^\infty \alpha z^{-\alpha-1}\delta_{\hat \boldb^{(j,k)}z}(\cdot)\,dz\,\int_{w_k^{-1}}^\infty\alpha w_k^{-\alpha} y^{-2\alpha-1}dy\,,\\
&=&
(c_\alpha^2 w_k^{2\alpha}/2)\sum_{j\ne k}\int_{0}^\infty \alpha z^{-\alpha-1}\delta_{\hat \boldb^{(j,k)}z}(\cdot)\,dz 
	=m_\alpha^{(k)}(\cdot)\,,
\end{eqnarray*}
proving \eqref{e:mean}.  

 Finally, the conditional independence of the extremes along with the laws of large numbers in \eqref{e:In.size} and \eqref{e:ratio.N12} show the required joint convergence along with the independence in the limit.
$\Box$

\bigskip

Below is a statement, needed in the sequel. A part of it was already used in the proof of Theorem \ref{pr:point_process}.

\begin{proposition}\label{prop:1} Under the assumptions of Theorem \ref{pr:point_process}, there is $C>0$ independent of $n$ and $k$ such that for all $z>0$, 
\begin{equation} \label{e:patt.i}
\bbP_{k,n}\left(u_nX^2_{i,-k}/Z_k^2>z\right) 
\leq C u_n^{-\alpha/2}z^{-\alpha/2}\,, 
\end{equation}
\begin{equation}  \label{e:patt.ii}
\bbP_{k,n}\left(u_n|T_{lk}|/Z_k>z\right) 
\leq C z^{-\alpha},
\end{equation}
\begin{equation} \label{e:patt.iii}
\E_{k,n}\left( |V_{l}|^{2\theta}\one\{|V_l|\le z\}\right)\le 
Cz^{2\theta-\alpha}\,.
\end{equation}

\end{proposition}

\medskip
\begin{proof}
 Since $|X_{i,-k}|\le a^*\sum_{j\ne k}Z_j$,  where $a^*=\max\{a_{ij}, i=1,\ldots,d; j=1,\ldots,p\}$, it follows that
\begin{equation*}
    \bbP(X_{i,-k}>z)\le \sum_{j\ne k}\bbP(a^* Z_j>z/p)\le Cz^{-\alpha}
\end{equation*}
for all $z$ and some $C>0$ by assumption \eqref{e:pareto}.  Thus,
\begin{eqnarray*}
        \bbP_{k,n}\left(u_nX^2_{i,-k}/Z_k^2>z\right)&\le& \frac{\bbP\left(X^2_{i,-k}>zu_n^{-1}Z_k^2, Z_k>u_n/w^{1/\alpha}\right)}
    {\bbP\left(\|\bX\|>u_n, Z_k>u_n/w^{1/\alpha}\right)}\\
   &\le& \frac{\bbP\left(X_{i,-k}>z^{1/2}u_n^{1/2}/w^{1/\alpha}\right) \bbP\left(Z_k>u_n/w^{1/\alpha}\right)}
    {\bbP\left(\|\bX\|>u_n, Z_k>u_n/w^{1/\alpha}\right)} \\
    &\le& Cu_n^{-\alpha/2}z^{-\alpha/2}\,,
\end{eqnarray*}
where $C$ may change value from line to line, and we have used the relation,
$\bbP(Z_k>u_n/w^{1/\alpha})/\bbP(\|\bX\|>u_n, Z_k>u_n/w^{1/\alpha})\to w_k^{-\alpha}$.
This proves \eqref{e:patt.i}. The same argument shows that 
\begin{eqnarray*}
        \bbP_{k,n}\left(u_n|T_{lk}|/Z_k>z\right)&\le& \frac{\bbP(|T_{lk}|>z/w^{1/\alpha})\bbP( Z_k>u_n/w^{1/\alpha})}
    {\bbP\left(\|\bX\|>u_n, Z_k>u_n/w^{1/\alpha}\right)}\\
       &\le& Cz^{-\alpha}\,,
\end{eqnarray*}
for all $z\ge 0$, proving \eqref{e:patt.ii}. Finally, it is straightforward to see from \eqref{e:patt.i}, \eqref{e:patt.ii} and \eqref{e:Vl} that
\begin{equation}
    \bbP_{k,n}(|V_l|>z)\le Cz^{-\alpha}\,,
\end{equation}
for some constant $C>0$. Therefore, 
\begin{eqnarray*}
    \E_{k,n}\left( |V_{l}|^{2\theta}\one\{|V_l|\le z\}\right)&\le& 2\theta \int_0^zu^{2\theta-1}\bbP_{k,n}(|V_l|>u)\,du\\
    &\le&C\int_0^zu^{2\theta-1}u^{-\alpha}\,du
    =Cz^{2\theta-\alpha}\,.
\end{eqnarray*}
 \end{proof}

\subsection*{Proof of Theorem \ref{thm:FrobScen2}}
					
 We start with the ``diagonal terms''. It follows from
\eqref{e:cov.zero}  and Theorem \ref{pr:Delta.B} that
\begin{align} \label{e:Fkk.power} 
F_{k,k}(n) =& d_\theta^2\frac{1}{N_n^2} \sum_{i\in \calI^{(k)}_n}
	\sum_{j\in \calI^{(k)}_n} \|
	\BY_i^{(k)}-\BY_j^{(k)}\|^{2\theta} + o_P(1).
\end{align}
By \eqref{e:In.size}, 
\begin{equation}\label{eq:fkk}
u_n^{4\theta-\alpha}n^{1-2\theta/\alpha}F_{k,k}(n)\sim \frac{d_\theta^2}{w^2c_\alpha^2}u_n^{\alpha}n^{-1}\int_{\bbr^d\times\bbr^d}\|\bx-\by\|^{2\theta}M_n^{(k)}(d\bx)\,M_n^{(k)}(d\by)\,.
\end{equation}
Notice that the scaling for $F_{kk}(n)$ is $u_n^{4\theta-\alpha}n^{1-2\theta/\alpha}=(u_n^2n^{-1/\alpha})^{2\theta}u_n^{-\alpha}n\to\infty$.

For $\vep>0$, set $B_\epsilon=\{\by: \|\by\|>\vep\}$. 	Take $0<\vep^\prime<\vep$, with $\vep^\prime$ much smaller than $\vep$.  By symmetry we can write 
\begin{eqnarray*}
\int_{\bbr^d\times\bbr^d}\|\bx-\by\|^{2\theta}M_n^{(k)}(d\bx)\,M_n^{(k)}(d\by)
&=&2\int_{B_\vep}\int_{B_{\vep^\prime}^c}+2\int_{B_\vep}\int_{B_{\vep^\prime}\cap B_\vep^c}+\int_{B_\vep}\int_{B_{\vep}}+\int_{B_\vep^c}\int_{B_\vep^c}\\
&=:& T_{\vep,\vep^\prime,n}^{(1)}+T_{\vep,\vep^\prime,n}^{(2)}+T_{\vep,n}^{(3)}+T_{\vep,n}^{(4)}\,.
\end{eqnarray*}
It will turn out that the asymptotic behaviour of $F_{kk}(n)$ will be determined by $T_{\vep,\vep^\prime,n}^{(1)}$.  To treat $ T_{\vep,\vep^\prime,n}^{(1)}$, we use a simple fact: if
$\|\boldb\|/\|\ba\|\leq \delta\in (0,1)$, then
$$
(1-\delta)^{2\theta}\|\ba\|^{2\theta}\leq \|\ba+\boldb\|^{2\theta}
\leq (1+\delta)^{2\theta}\|\ba\|^{2\theta}.
$$
By taking $\delta=\vep^\prime/\vep$, we obtain
\begin{equation} \label{e:Teps.1simp}
u_n^{\alpha}n^{-1}	T_{\vep,\vep^\prime,n}^{(1)} \in \Bigl[ \bigl(
	1-\vep^\prime/\vep\bigr)^{2\theta}, 
	\bigl(
	1+\vep^\prime/\vep\bigr)^{2\theta}\Bigr]\, 
	u_n^{\alpha}n^{-1} 2\int_{B_\vep}\|\by\|^{2\theta}\left(\int_{B_{\vep^\prime}^c}M_n^{(k)}(d\bx)\right)\,M_n^{(k)}(d\by)\,.
\end{equation}
Since $u_n^{\alpha}n^{-1}M_n^{(k)}(B_{\vep^\prime}^c)=u_n^{\alpha}n^{-1}(N_n^{(k)}-M_n^{(k)}(B_{\vep^\prime}))$ and $M_n^{(k)}(B_{\vep^\prime})\Rightarrow  M_\alpha^{(k)}(B_{\vep^\prime})<\infty$ a.s., we conclude that 
\begin{eqnarray}
	u_n^{\alpha}n^{-1} 2\int_{B_\vep}\|\by\|^{2\theta}\left(\int_{B_{\vep^\prime}^c}M_n^{(k)}(d\bx)\right)\,M_n^{(k)}(d\by)
	&=&2\int_{B_\vep}\|\by\|^{2\theta}M_n^{(k)}(d\by)\,u_n^{\alpha}n^{-1}M_n^{(k)}(B_{\vep^\prime}^c)\nonumber\\
	&\Rightarrow & 2c_\alpha w_k^\alpha\int_{\|\by\|>\vep}\|\by\|^{2\theta}M_\alpha^{(k)}(d\by)\,.\label{eq:fkklimit1}
\end{eqnarray}
From the weak convergence in \eqref{e:pp.conv}, it follows directly that
\begin{equation*}
	2\int_{B_\vep}\int_{B_{\vep^\prime}\cap B_\vep^c}\|\bx-\by\|^{2\theta}M_n^{(k)}(d\bx)\,M_n^{(k)}(d\by)\Rightarrow
		2\int_{\|\by\|>\vep}\int_{\vep^\prime<\|\bx\|\le \vep}\|\bx-\by\|^{2\theta}M_\alpha^{(k)}(d\bx)\,M_\alpha^{(k)}(d\by)
\end{equation*}
and hence
\begin{equation*} 
	u_n^{\alpha}n^{-1}T_{\vep,\vep^\prime,n}^{(2)}\cip 0\,.
\end{equation*}
Combining this result with \eqref{eq:fkklimit1}, we obtain,
\begin{equation}\label{eq:t2}
	u_n^{\alpha}n^{-1}(T_{\vep,\vep^\prime,n}^{(1)}+T_{\vep,\vep^\prime,n}^{(2)})\Rightarrow 2c_\alpha w_k^\alpha\int_{\|\by\|>\vep}\|\by\|^{2\theta}M_\alpha^{(k)}(d\by)\,.
\end{equation}
Turning to $T_{\vep,n}^{(3)}$, we have once again from the point process convergence in \eqref{e:pp.conv},
					\begin{equation*}
					T_{\vep,n}^{(3)}= \int_{B_\vep}\left(\int_{B_\vep}\|\by-\bx\|^{2\theta}M_n^{(k)}(d\bx)\right)\,M_n^{(k)}(d\by)
				\Rightarrow \int_{B_\vep}\left(\int_{B_\vep}\|\by-\bx\|^{2\theta}M_\alpha^{(k)}(d\bx)\right)\,M_\alpha^{(k)}(d\by)\,,
					\end{equation*}
from which we conclude,
\begin{equation}\label{e:smallT.eps}
	u_n^{\alpha}n^{-1}T_{\vep,n}^{(3)}\cip 0\,.
\end{equation}
Finally we handle the last term $T_{\vep,n}^{(4)}$. We have
\begin{equation} \label{e:Teps.2a}
	u_n^{\alpha}n^{-1}T_{\vep,n}^{(4)}\le 2\int_{B_\vep}\|\by\|^{2\theta}M_n^{(k)}(d\by)\left(u_n^{\alpha}n^{-1}\right)N_n^{(k)}
\end{equation}
and, since $u_n^{-\alpha} n^{-1}N_n^{(k)}\cip c_\alpha w_k^\alpha$, we restrict attention to the integral.  By \eqref{e:patt.iii} 
with $z=\epsilon u_n^{-1}n^{1/\alpha}$,  
\begin{eqnarray*}
	\E\left( \int_{B_\vep}\|\by\|^{2\theta}M_n^{(k)}(d\by)\right)&\sim& w_k^\alpha c_\alpha u_n^{2\theta-\alpha}n^{1-2\theta/\alpha} \E_{k,n}\left(\|\BV\|^{2\theta}\one(\|\BV\|\le  \epsilon u_n^{-1}n^{1/\alpha})\right)\\
	&\le&C\epsilon^{2\theta-\alpha}\,,
\end{eqnarray*}
and hence
\begin{equation*}
\lim_{\epsilon\to 0}\limsup_{n\to\infty}\E\left( \int_{B_\vep}\|\by\|^{2\theta}M_n^{(k)}(d\by)\right)=0\,.
\end{equation*}
This, in turn, implies from \eqref{e:Teps.2a} that for any $\eta>0$,
\begin{equation} \label{eq:t4}
\lim_{\epsilon\to 0}\limsup_{n\to\infty}\bbP(u_n^{\alpha}n^{-1}T_{\vep,n}^{(4)}>\eta)=0.
\end{equation}
Combining \eqref{eq:fkklimit1}, \eqref{eq:t2}, \eqref{e:smallT.eps}, and \eqref{eq:t4}, we obtain
\begin{equation}\label{eq:fkk3}
u_n^{4\theta-\alpha}n^{1-2\theta/\alpha}F_{k,k}(n)\Rightarrow \frac{2d_\theta^2w_k^\alpha}{w^2c_\alpha}\int_{\|\by\|>0}\|\by\|^{2\theta}M_\alpha^{(k)}(d\by)\,.
\end{equation}

The ``off-diagonal terms'' can be treated in a similar fashion. It follows
from the continuous differentiability of $R$ outside the origin and
Theorem \ref{pr:Delta.B} that, in the obvious notation, 
\begin{align} \label{e:Fk1k2.power}
F_{k_1,k_2}(n) =& \frac{1}{N_n^2}\sum_{i=1}^{N_n^{(k_1)}} \sum_{j=1}^{N_n^{(k_2)}} \Bigl[ 
             R\bigl( \BY_i^{(k_1)}-\BY_j^{(k_2)}\bigr)
                  - R\bigl( \bs_{k_1}-\bs_{k_2}\bigr)\Bigr]^2 \\
\notag =& \frac{1}{N_n^2} \sum_{i=1}^{N_n^{(k_1)}} \sum_{j=1}^{N_n^{(k_2)}}
\Bigl[ \big\langle \nabla
R(\bs_{k_1}-\bs_{k_2}),  (\BY_i^{(k_1)}-\bs_{k_1})-(\BY_j^{(k_2)}-\bs_{k_2})\big\rangle\Bigr]^2 + o_P(1)\,.
\end{align}
Arguing as before (see \eqref{eq:fkk}),
\begin{equation*}
u_n^{4-\alpha}n^{1-2/\alpha}F_{k_1,k_2}(n)\sim \frac{1}{w^2c_\alpha^2}u_n^{\alpha}n^{-1}\int_{\bbr^d\times\bbr^d}
\left[\langle\nabla
R(\bs_{k_1}-\bs_{k_2}),  \bx-\by\rangle\right]^2
M_n^{(k_1)}(d\bx)\,M_n^{(k_2)}(d\by)\,.
\end{equation*}
We the same notation, as above we write 
\begin{eqnarray*}
\int_{\bbr^d\times\bbr^d} &=&
\int_{B_\vep}\int_{B_{\vep^\prime}^c}+\int_{B_\vep}\int_{B_{\vep^\prime}\cap B_\vep^c}+\int_{B_{\vep^\prime}^c}\int_{B_\vep}+\int_{B_{\vep^\prime}\cap B_\vep^c}\int_{B_\vep} 
 +\int_{B_\vep}\int_{B_{\vep}}+\int_{B_\vep^c}\int_{B_\vep^c}\\
&=:& T_{\vep,\vep^\prime,n}^{(1,1)}+T_{\vep,\vep^\prime,n}^{(1,2)}+T_{\vep,\vep^\prime,n}^{(2,1)}+T_{\vep,\vep^\prime,n}^{(2,2)}+T_{\vep,n}^{(3)}+T_{\vep,n}^{(4)}\,.
\end{eqnarray*}
These six terms are treated in  nearly the same way as earlier for the  analogous decomposition for $F_{kk}(n)$.  In this vein, we have for the first term
\begin{equation} \label{e:approx.T11}
u_n^\alpha n^{-1} T_{\vep,\vep^\prime,n}^{(1,1)}\in \Bigl[ \bigl(
  1-\vep^\prime/\vep\bigr)^2, 
\bigl(
  1+\vep^\prime/\vep\bigr)^{2}\Bigr]u_n^{\alpha}n^{-1}\int_{B_{\vep}}\left[\bigl\langle \nabla
     R(\bs_{k_1}-\bs_{k_2}), \bx\rangle\right]^2 M_n^{(k_1)}(d\bx)M_n^{(k_2)}(B_{\vep^\prime}^c)\,.
\end{equation}
Since $u_n^{\alpha}n^{-1}M_n^{(k_2)}(B_{\vep^\prime}^c)\sim u_n^{\alpha}n^{-1}N_n^{(k_2)}\cip c_\alpha w_{k_2}^\alpha$, we have  by the weak convergence in \eqref{e:pp.conv}
\begin{align}
\notag &u_n^{\alpha}n^{-1}\int_{B_{\vep}}\left[\langle \nabla
     R(\bs_{k_1}-\bs_{k_2}), \bx\rangle\right]^2 M_n^{(k_1)}(d\bx)M_n^{(k_2)}(B_{\vep^\prime}^c) \nonumber\\
   &~~~~ \Rightarrow 
     c_\alpha w_{k_2}^\alpha \int_{B_{\vep}}\left[\langle \nabla
     R(\bs_{k_1}-\bs_{k_2}), \bx\rangle\right]^2 M_\alpha^{(k_1)}(d\bx)\,.\label{eq:t11}
\end{align}
Next, a straightforward application of the weak convergence in \eqref{e:pp.conv}, gives
\begin{align*}
	&\int_{B_\vep}\left(\int_{B_{\vep^\prime}\cap B_\vep^c}\left[\langle\nabla
R(\bs_{k_1}-\bs_{k_2}),  \bx-\by\rangle\right]^2M_n^{(k_2)}(d\by)\right)\,M_n^{(k_1)}(d\bx)\\
&~~~~\Rightarrow
		\int_{\|\bx\|>\vep}\left(\int_{\vep^\prime<\|\by\|\le \vep}\left[\langle\nabla
R(\bs_{k_1}-\bs_{k_2}),  \bx-\by\rangle\right]^2M_\alpha^{(k_2)}(d\by)\right)\,M_\alpha^{(k_1)}(d\bx)
\end{align*}
and hence
\begin{equation}\label{eq:t2a}
	u_n^{\alpha}n^{-1}T_{\vep,\vep^\prime,n}^{(1,2)}\cip 0\,.
\end{equation}
Interchanging the roles of $\bx$ and $\by$, the terms $T_{\vep,\vep^\prime,n}^{(1,2)}$ and $T_{\vep,\vep^\prime,n}^{(2,2)}$ are handled in exactly the same way.  The same reasoning (see \eqref{e:smallT.eps})  also shows that 
\begin{equation} \label{eq:t3}
   u_n^{-\alpha}n^{-1} T_{\vep,n}^{(3)}\cip 0.
\end{equation}
Finally, turning to $T_{\vep,n}^{(4)}$, we have by the  Cauchy-Schwarz inequality, 
\begin{equation*} 
	u_n^{\alpha}n^{-1}T_{\vep,n}^{(4)}\le C\left(\int_{B_\vep}\|\by\|^{2}M_n^{(k_1)}(d\by)\left(u_n^{\alpha}n^{-1}\right)N_n^{(k_2)}+\int_{B_\vep}\|\by\|^{2}M_n^{(k_2)}(d\by)\left(u_n^{\alpha}n^{-1}\right)N_n^{(k_1)}\right)
\end{equation*}
for some constant $C>0$.  But now the exact same argument leading to \eqref{eq:t4} (with $\theta=1$) can be applied to each of the summands from which it follows that for any $\eta>0$
\begin{equation} \label{eq:t4a}
\lim_{\epsilon\to 0}\limsup_{n\to\infty}\bbP(u_n^{\alpha}n^{-1}T_{\vep,n}^{(4)}>\eta)=0\,.
\end{equation}

Combining the results in \eqref{e:approx.T11}, \eqref{eq:t11}, \eqref{eq:t2a}, \eqref{eq:t3}, and \eqref{eq:t4a}, we conclude that 
\begin{equation} \label{e:offd.lim}
\bigl(  u_n^{4-\alpha}n^{1-2/\alpha}\bigr) 
F_{k_1,k_2}(n) 
\Rightarrow \frac{1}{w^2c_\alpha} \int_{\bbr^d} \Bigl[\langle \nabla
R(\bs_{k_1}-\bs_{k_2}),  \bx\rangle\Bigr]^2 \,
                     \bigl( w_{k_2}^2M_\alpha^{(k_1)}+  w_{k_1}^2M_\alpha^{(k_2)}\bigr)
(d\bx).
\end{equation}
By Theorem \ref{pr:point_process}  the convergence in \eqref{e:offd.lim}  and \eqref{eq:fkk3} is joint in $k_1,k_2$, 
and the claim of the theorem follows. 
Note that if $\theta=1$, then the scaling for $F_{k_1,k_2}(n)$ is the same as that for $F_{kk}$. On the other hand if $\theta>1$, then $\bigl(u_n^{4-\alpha}n^{1-2/\alpha}\bigr) 
F_{k_1,k_2}(n)\cip 0$ so that the diagonal terms are of smaller order.     
$\Box$

\subsection*{Proof of Theorem \ref{thm:FrobScen.mid}}

In this case the order of magnitude of the ``diagonal terms'' is still
given by \eqref{e:kk.scen1a}  while the order of magnitude of the
``off-diagonal terms'' is given by \eqref{e:k1k2.scen1f}, because the
latter statement requires only that $\alpha>2$. We claim that 
\begin{equation} \label{e:compare.terms}
u_n^{-4\theta+\alpha}n^{2\theta/\alpha-1}\ll u_n^{-2}.
\end{equation} 
Indeed, \eqref{e:compare.terms} is equivalent to
$$
n\ll u_n^{(4\theta-\alpha-2)/(2\theta/\alpha-1)},
$$
which is a true statement due to \eqref{e:level.unusual} and the fact
that
$$
\frac{4\theta-\alpha-2}{2\theta/\alpha-1}>2\alpha
$$
as implied by the condition $2<\alpha<2\theta$.  It follows from \eqref{e:compare.terms} that the ``off-diagonal
terms'' are of a larger order of magnitude than the ``diagonal
terms''.  
$\Box$

\bibliographystyle{plainnat}
\bibliography{biblio}

\begin{thebibliography}{39}
\providecommand{\natexlab}[1]{#1}
\providecommand{\url}[1]{\texttt{#1}}
\expandafter\ifx\csname urlstyle\endcsname\relax
  \providecommand{\doi}[1]{doi: #1}\else
  \providecommand{\doi}{doi: \begingroup \urlstyle{rm}\Url}\fi

\bibitem[Avella~Medina et~al.(2021)Avella~Medina, Davis, and
  Samorodnitsky]{avellamedinaetal2021}
Marco Avella~Medina, Richard~A Davis, and Gennady Samorodnitsky.
\newblock Spectral learning of multivariate extremes.
\newblock \emph{arXiv preprint arXiv:2111.07799}, 2021.

\bibitem[Bak{\i}r et~al.(2004)Bak{\i}r, Weston, and
  Sch{\"o}lkopf]{bakiretal2004}
G{\"o}khan~H Bak{\i}r, Jason Weston, and Bernhard Sch{\"o}lkopf.
\newblock Learning to find pre-images.
\newblock \emph{Advances in Neural Information Processing Systems},
  16:\penalty0 449--456, 2004.

\bibitem[Basrak et~al.(2002{\natexlab{a}})Basrak, Davis, and
  Mikosch]{basrak2002characterization}
Bojan Basrak, Richard~A. Davis, and Thomas Mikosch.
\newblock A characterization of multivariate regular variation.
\newblock \emph{Annals of Applied Probability}, 12\penalty0 (3):\penalty0
  908--920, 2002{\natexlab{a}}.

\bibitem[Basrak et~al.(2002{\natexlab{b}})Basrak, Davis, and
  Mikosch]{basrak2002regular}
Bojan Basrak, Richard~A. Davis, and Thomas Mikosch.
\newblock Regular variation of {GARCH} processes.
\newblock \emph{Stochastic Processes and their Applications}, 99\penalty0
  (1):\penalty0 95--115, 2002{\natexlab{b}}.

\bibitem[Blanchard et~al.(2007)Blanchard, Bousquet, and
  Zwald]{blanchardetal2007}
Gilles Blanchard, Olivier Bousquet, and Laurent Zwald.
\newblock Statistical properties of kernel principal component analysis.
\newblock \emph{Machine Learning}, 66\penalty0 (2):\penalty0 259--294, 2007.

\bibitem[Braun et~al.(2008)Braun, Buhmann, and M{\"u}ller]{braunetal2008}
Mikio~L Braun, Joachim~M Buhmann, and Klaus-Robert M{\"u}ller.
\newblock On relevant dimensions in kernel feature spaces.
\newblock \emph{Journal of Machine Learning Research}, 9:\penalty0 1875--1908,
  2008.

\bibitem[Chautru(2015)]{chautru2015}
Emilie Chautru.
\newblock Dimension reduction in multivariate extreme value analysis.
\newblock \emph{Electronic Journal of Statistics}, 9\penalty0 (1):\penalty0
  383--418, 2015.

\bibitem[Cl{\'e}men{\c{c}}on et~al.(2021)Cl{\'e}men{\c{c}}on, Jalalzai,
  Sabourin, and Segers]{clemencconetal2021}
St{\'e}phan Cl{\'e}men{\c{c}}on, Hamid Jalalzai, Anne Sabourin, and Johan
  Segers.
\newblock Concentration bounds for the empirical angular measure with
  statistical learning applications.
\newblock \emph{arXiv preprint arXiv:2104.03966}, 2021.

\bibitem[Cooley and Thibaud(2019)]{cooleyandthibaud2019}
Daniel Cooley and Emeric Thibaud.
\newblock Decompositions of dependence for high-dimensional extremes.
\newblock \emph{Biometrika}, 106\penalty0 (3):\penalty0 587--604, 2019.

\bibitem[Deuber et~al.(2021)Deuber, Li, Engelke, and Maathuis]{deuberetal2021}
David Deuber, Jinzhou Li, Sebastian Engelke, and Marloes~H Maathuis.
\newblock Estimation and inference of extremal quantile treatment effects for
  heavy-tailed distributions.
\newblock \emph{arXiv preprint arXiv:2110.06627}, 2021.

\bibitem[Drees and Sabourin(2021)]{dreesandsabourin2021}
Holger Drees and Anne Sabourin.
\newblock Principal component analysis for multivariate extremes.
\newblock \emph{Electronic Journal of Statistics}, 15\penalty0 (1):\penalty0
  908--943, 2021.

\bibitem[Engelke and Hitz(2020)]{engelkeandhitz2020}
Sebastian Engelke and Adrien~S Hitz.
\newblock Graphical models for extremes.
\newblock \emph{Journal of the Royal Statistical Society: Series B},
  82\penalty0 (4):\penalty0 871--932, 2020.

\bibitem[Engelke and Ivanovs(2021)]{engelkeandivanovs2021}
Sebastian Engelke and Jevgenijs Ivanovs.
\newblock Sparse structures for multivariate extremes.
\newblock \emph{Annual Review of Statistics and Its Application}, 8:\penalty0
  241--270, 2021.

\bibitem[Engelke et~al.(2022)Engelke, Lalancette, and
  Volgushev]{engelkeetal2022}
Sebastian Engelke, Micha{\"e}l Lalancette, and Stanislav Volgushev.
\newblock Learning extremal graphical models in high dimensions.
\newblock \emph{arXiv preprint arXiv:2111.00840}, 2022.

\bibitem[Fomichov and Ivanovs(2022)]{fomichovandivanovs2022}
V~Fomichov and J~Ivanovs.
\newblock Spherical clustering in detection of groups of concomitant extremes.
\newblock \emph{Biometrika}, 2022.

\bibitem[Gnecco et~al.(2021)Gnecco, Meinshausen, Peters, and
  Engelke]{gneccoetal2021}
Nicola Gnecco, Nicolai Meinshausen, Jonas Peters, and Sebastian Engelke.
\newblock Causal discovery in heavy-tailed models.
\newblock \emph{Annals of Statistics}, 49\penalty0 (3):\penalty0 1755--1778,
  2021.

\bibitem[Goix et~al.(2015)Goix, Sabourin, and Cl{\'e}men]{goixetal2015}
Nicolas Goix, Anne Sabourin, and St{\'e}phan Cl{\'e}men.
\newblock Learning the dependence structure of rare events: a non-asymptotic
  study.
\newblock In \emph{Conference on Learning Theory}, pages 843--860. PMLR, 2015.

\bibitem[Goix et~al.(2017)Goix, Sabourin, and
  Cl{\'e}men{\c{c}}on]{goixetal2017}
Nicolas Goix, Anne Sabourin, and Stephan Cl{\'e}men{\c{c}}on.
\newblock Sparse representation of multivariate extremes with applications to
  anomaly detection.
\newblock \emph{Journal of Multivariate Analysis}, 161:\penalty0 12--31, 2017.

\bibitem[Honeine and Richard(2011)]{honeineandcedric2011}
Paul Honeine and Cedric Richard.
\newblock Preimage problem in kernel-based machine learning.
\newblock \emph{IEEE Signal Processing Magazine}, 28\penalty0 (2):\penalty0
  77--88, 2011.

\bibitem[Jalalzai and Leluc(2021)]{jalalzaiandleluc2021}
Hamid Jalalzai and R{\'e}mi Leluc.
\newblock Feature clustering for support identification in extreme regions.
\newblock In \emph{International Conference on Machine Learning}, pages
  4733--4743. PMLR, 2021.

\bibitem[Jan{\ss}en and Wan(2020)]{janssenandwan2020}
Anja Jan{\ss}en and Phyllis Wan.
\newblock $ k $-means clustering of extremes.
\newblock \emph{Electronic Journal of Statistics}, 14\penalty0 (1):\penalty0
  1211--1233, 2020.

\bibitem[Kim et~al.(2002)Kim, Jung, and Kim]{kimetal2002}
Kwang~In Kim, Keechul Jung, and Hang~Joon Kim.
\newblock Face recognition using kernel principal component analysis.
\newblock \emph{IEEE signal processing letters}, 9\penalty0 (2):\penalty0
  40--42, 2002.

\bibitem[Kwok and Tsang(2004)]{kwokandtsang2004}
James~T.Y. Kwok and Ivor~W.H Tsang.
\newblock The pre-image problem in kernel methods.
\newblock \emph{IEEE Transactions on Neural Networks}, 15\penalty0
  (6):\penalty0 1517--1525, 2004.

\bibitem[Meyer and Wintenberger(2019)]{meyerandwintenberger2019}
Nicolas Meyer and Olivier Wintenberger.
\newblock Sparse regular variation.
\newblock \emph{arXiv preprint arXiv:1907.00686}, 2019.

\bibitem[Mika et~al.(1998)Mika, Sch{\"o}lkopf, Smola, M{\"u}ller, Scholz, and
  R{\"a}tsch]{mikaetal1998}
Sebastian Mika, Bernhard Sch{\"o}lkopf, Alex Smola, Klaus-Robert M{\"u}ller,
  Matthias Scholz, and Gunnar R{\"a}tsch.
\newblock Kernel {PCA} and de-noising in feature spaces.
\newblock \emph{Advances in Neural Information Processing Systems}, 11, 1998.

\bibitem[Resnick(2007)]{resnick2007}
Sidney~I. Resnick.
\newblock \emph{Heavy-tail phenomena: probabilistic and statistical modeling}.
\newblock Springer Science \& Business Media, 2007.

\bibitem[Resnick(2008)]{resnick2008}
Sidney~I. Resnick.
\newblock \emph{Extreme values, regular variation, and point processes},
  volume~4.
\newblock Springer Science \& Business Media, 2008.

\bibitem[Rohrbeck and Cooley(2022)]{rohrbeckandcooley2022}
Christian Rohrbeck and Daniel Cooley.
\newblock Simulating flood event sets using extremal principal components.
\newblock \emph{Annals of Applied Statistics (to appear)}, 2022.

\bibitem[Rosipal et~al.(2001)Rosipal, Girolami, Trejo, and
  Cichocki]{rosipaletal2001}
Roman Rosipal, Mark Girolami, Leonard~J Trejo, and Andrzej Cichocki.
\newblock Kernel {PCA} for feature extraction and de-noising in nonlinear
  regression.
\newblock \emph{Neural Computing \& Applications}, 10\penalty0 (3):\penalty0
  231--243, 2001.

\bibitem[Sch{\"o}lkopf et~al.(1997)Sch{\"o}lkopf, Smola, and
  M{\"u}ller]{scholkopfetal1997}
Bernhard Sch{\"o}lkopf, Alexander Smola, and Klaus-Robert M{\"u}ller.
\newblock Kernel principal component analysis.
\newblock In \emph{International Conference on Artificial Neural Networks},
  pages 583--588. Springer, 1997.

\bibitem[Shawe-Taylor et~al.(2005)Shawe-Taylor, Williams, Cristianini, and
  Kandola]{shawetayloretal2005}
John Shawe-Taylor, Christopher~KI Williams, Nello Cristianini, and Jaz Kandola.
\newblock On the eigenspectrum of the gram matrix and the generalization error
  of kernel-{PCA}.
\newblock \emph{IEEE Transactions on Information Theory}, 51\penalty0
  (7):\penalty0 2510--2522, 2005.

\bibitem[Simpson et~al.(2020)Simpson, Wadsworth, and Tawn]{simpsonetal2020}
Emma~S. Simpson, Jennifer~L. Wadsworth, and Jonathan~A. Tawn.
\newblock Determining the dependence structure of multivariate extremes.
\newblock \emph{Biometrika}, 107\penalty0 (3):\penalty0 513--532, 2020.

\bibitem[Smola and Sch{\"o}lkopf(1998)]{smolaandscholkopt1998}
Alex~J. Smola and Bernhard Sch{\"o}lkopf.
\newblock \emph{Learning with Kernels}, volume~4.
\newblock Citeseer, 1998.

\bibitem[Steinwart and Christmann(2008)]{steinwartandchristmann2008}
Ingo Steinwart and Andreas Christmann.
\newblock \emph{Support Vector Machines}.
\newblock Springer Science \& Business Media, 2008.

\bibitem[Tyler(1987)]{tyler1987}
David~E. Tyler.
\newblock Statistical analysis for the angular central {G}aussian distribution
  on the sphere.
\newblock \emph{Biometrika}, 74\penalty0 (3):\penalty0 579--589, 1987.

\bibitem[van Zanten and van~der Vaart(2008)]{vanzantenandvandervaart2008}
J.~Harry van Zanten and Aad~W. van~der Vaart.
\newblock Reproducing kernel {H}ilbert spaces of {G}aussian priors.
\newblock In \emph{Pushing the limits of contemporary statistics: contributions
  in honor of Jayanta K. Ghosh}, pages 200--222. Institute of Mathematical
  Statistics, 2008.

\bibitem[Yu et~al.(2015)Yu, Wang, and Samworth]{yuetal2015}
Yi~Yu, Tengyao Wang, and Richard~J Samworth.
\newblock A useful variant of the {D}avis--{K}ahan theorem for statisticians.
\newblock \emph{Biometrika}, 102\penalty0 (2):\penalty0 315--323, 2015.

\bibitem[Zheng et~al.(2010)Zheng, Lai, and Yuen]{zhengetal2010}
Wei-Shi Zheng, JianHuang Lai, and Pong~C Yuen.
\newblock Penalized preimage learning in kernel principal component analysis.
\newblock \emph{IEEE Transactions on Neural Networks}, 21\penalty0
  (4):\penalty0 551--570, 2010.

\bibitem[Zwald and Blanchard(2005)]{zwaldandblanchard2005}
Laurent Zwald and Gilles Blanchard.
\newblock On the convergence of eigenspaces in kernel principal component
  analysis.
\newblock \emph{Advances in Neural Information Processing Systems}, 18, 2005.

\end{thebibliography}

\end{document}